\documentclass[format=acmsmall, screen=true, review=false, anonymous=false]{acmart}
\AtBeginDocument{%
  }

\setcopyright{acmlicensed}
\copyrightyear{2025}
\acmYear{2025}
\acmDOI{XXXXXXX.XXXXXXX}

\usepackage{subfloat}
\usepackage{subcaption}
\usepackage{mathtools}

\newcommand{\E}{\mathbb{E}}

\DeclareMathOperator*{\argmin}{\arg\!\min}

\begin{document}

\title{Optimal Resource Allocation for ML Model Training and Deployment under Concept Drift}

\author{Hasan Burhan Beytur}
\affiliation{%
  \institution{The University of Texas at Austin}
  \city{Austin}
  \state{Texas}
  \country{USA}}
\email{hbbeytur@utexas.edu}

\author{Haris Vikalo}
\affiliation{%
  \institution{The University of Texas at Austin}
  \city{Austin}
  \state{Texas}
  \country{USA}}
\email{hvikalo@ece.utexas.edu}

\author{Kevin S Chan}
\affiliation{%
  \institution{Devcom Army Research Laboratory}
  \country{USA}}
\email{kevin.s.chan.civ@army.mil}

\author{Gustavo de Veciana}
\affiliation{%
  \institution{The University of Texas at Austin}
  \city{Austin}
  \state{Texas}
  \country{USA}}
\email{deveciana@utexas.edu}

% \renewcommand{\shortauthors}{Beytur et al.}

\begin{abstract}
We study how to allocate resources for training and deployment of machine learning (ML) models under concept drift and limited budgets. We consider a setting in which a model provider distributes trained models to multiple clients whose devices support local inference but lack the ability to retrain those models, placing the burden of performance maintenance on the provider. We introduce a model-agnostic framework that captures the interaction between resource allocation, concept drift dynamics, and deployment timing. We show that optimal training policies depend critically on the aging properties of concept durations. Under sudden concept changes, we derive optimal training policies subject to budget constraints when concept durations follow distributions with Decreasing Mean Residual Life (DMRL), and show that intuitive heuristics are provably suboptimal under Increasing Mean Residual Life (IMRL). We further study model deployment under communication constraints, prove that the associated optimization problem is quasi-convex under mild conditions, and propose a randomized scheduling strategy that achieves near-optimal client-side performance. These results offer theoretical and algorithmic foundations for cost-efficient ML model management under concept drift, with implications for continual learning, distributed inference, and adaptive ML systems.

% These results provide theoretical foundations and algorithmic principles for designing resilient ML systems.
\end{abstract}

\begin{CCSXML}
<ccs2012>
 <concept>
  <concept_id>00000000.0000000.0000000</concept_id>
  <concept_desc>Do Not Use This Code, Generate the Correct Terms for Your Paper</concept_desc>
  <concept_significance>500</concept_significance>
 </concept>
 <concept>
  <concept_id>00000000.00000000.00000000</concept_id>
  <concept_desc>Do Not Use This Code, Generate the Correct Terms for Your Paper</concept_desc>
  <concept_significance>300</concept_significance>
 </concept>
 <concept>
  <concept_id>00000000.00000000.00000000</concept_id>
  <concept_desc>Do Not Use This Code, Generate the Correct Terms for Your Paper</concept_desc>
  <concept_significance>100</concept_significance>
 </concept>
 <concept>
  <concept_id>00000000.00000000.00000000</concept_id>
  <concept_desc>Do Not Use This Code, Generate the Correct Terms for Your Paper</concept_desc>
  <concept_significance>100</concept_significance>
 </concept>
</ccs2012>
\end{CCSXML}

\ccsdesc[500]{Do Not Use This Code~Generate the Correct Terms for Your Paper}
\ccsdesc[300]{Do Not Use This Code~Generate the Correct Terms for Your Paper}
\ccsdesc{Do Not Use This Code~Generate the Correct Terms for Your Paper}
\ccsdesc[100]{Do Not Use This Code~Generate the Correct Terms for Your Paper}

\keywords{Concept Drift, Resource Allocation, Model Training, Model Deployment, MLOps, Optimal Control, Functional Optimization}


\maketitle

\section{Introduction}

The rapid advancement of machine learning (ML), particularly in generative image and large language models (LLMs), is transforming how we carry out both professional and personal tasks
\cite{linkonAdvancementsApplicationsGenerative2024,kimToolsUnderstandingHow2025,husseinChatGPTsImpactSectors2025}. As such models grow in complexity, their training and inference require increasingly more data and computational power, rendering their management resource-intensive. While these demands have largely been met by centralized cloud infrastructure, there is a notable shift toward localized inference on end-user devices. This transition improves scalability by reducing reliance on shared resources and enhances user privacy
\cite{park_llamaduo_2024,zhou_tinyllava_2024,chenWhatRoleSmall2024}. However, despite the rise of on-device inference, training remains predominantly cloud-based due to its substantial computational and coordination costs.

In real-world applications, ML model performance often degrades over time due to shifts in underlying data distributions -- a phenomenon known as \textit{concept drift}. Such drift may arise from changes in the relationship between input features and target variables, evolving class distributions, or the emergence of out-of-distribution data. For instance, a rental price predictor may lose accuracy as market conditions evolve; a clinical model may degrade due to demographic changes; and an LLM may fail to answer questions about recent events. In all cases, performance degradation stems from a growing mismatch between training and deployment data \cite{gama_survey_2014, han_survey_2022, lu_learning_2019}. Maintaining performance under concept drift requires continuous monitoring, retraining, and redeployment -- an operational challenge addressed by the field of MLOps \cite{kreuzberger2022machinelearningoperationsmlops, stone2025navigating,muravevMLOpsArchitectureFuture}. These tasks add considerable overhead to an already costly model development cycle. As ML adoption grows and models become more complex, efficient resource allocation within MLOps workflows becomes critical for scalable and sustainable deployment.

We consider a setting in which a service provider trains ML models and distributes them to client devices for local inference. While clients are capable of performing inference, they often lack the computational and data resources needed for retraining. As a result, the responsibility for maintaining model accuracy under concept drift falls on the provider, who must judiciously allocate limited resources for retraining and model updates as performance degrades. We focus on scenarios involving sudden, discrete drift events and introduce a model-agnostic framework that captures how performance depends jointly on training resource allocation and the dynamics of concept change. Since client-side utility hinges on timely model refresh, we further study drift-aware deployment strategies aimed at optimizing long-term average performance.
Our results offer practical guidance for designing resilient, resource-efficient ML systems that adapt to evolving data.

\subsection{Related Works}

Extensive empirical studies have investigated the trade-offs among compute, model size, and data volume in ML training. A foundational work in this area \cite{kaplan_scaling_2020} demonstrated that the cross-entropy loss in autoregressive language models follows power-law relationships with model size, dataset size, and total compute. These scaling laws, observed across several orders of magnitude, offer simple prescriptions for allocating fixed compute budgets to achieve target performance. Building on this, \cite{hoffmann_training_2022} showed that language models are most compute-efficient when model size and training token count are scaled proportionally. This insight led to the development of Chinchilla (70B parameters, 1.4T tokens), which outperforms much larger models under the same compute constraints. Complementary work on learning curve modeling has enabled more cost-effective training via early stopping and budget-aware learning rate schedules \cite{domhan_speedingup_2015,Viering_2023_Shape,li2019budgeted}. For example, \cite{domhan_speedingup_2015} proposed a probabilistic model to extrapolate learning curves and terminate unpromising hyperparameter configurations early. Similarly, \cite{li2019budgeted} provided empirical evidence that budget-aware learning rate schedules can improve efficiency of training in resource-constrained settings. Inspired by these insights into training-time efficiency, our work extends this line of research to include deployment-time considerations in nonstationary environments. Specifically, we study how training resource allocation and model deployment strategies affect performance under sudden concept drift -- an important yet underexplored regime in the context of scaling laws and resource-aware training.

Concept drift challenges the standard assumption of stationary data distributions in ML. Seminal surveys classify drift as either gradual (incremental) or sudden (abrupt) changes in data distribution \cite{gama_survey_2014, lu_learning_2019}. Gradual drift causes slow, predictable performance degradation and is commonly addressed through sliding-window retraining \cite{baier2020switchingschemenovelapproach}, instance reweighting \cite{kolter_lessons_2007}, or time-varying optimization frameworks \cite{mokhtari_online_2016, simonetto_class_2016}. For example, \cite{simonetto_class_2016} introduced a two-stage online algorithm that tracks a time-varying optimum, providing theoretical bounds that link drift rate to the number of required optimization steps—and, by extension, to resource demand. In contrast, sudden concept drift, characterized by its abrupt and unpredictable nature, typically requires immediate model replacement or resets triggered by change-point detection \cite{bifetLearningTimeChangingData2007,budka_change-point_2018,moreno-gracia_dynamic_2020,mahadevan2023cost,vzliobaite2015towards,regol2025when}. Due to their distinct dynamics, gradual and sudden drift require different mitigation strategies \cite{han_survey_2022,lu_learning_2019}. While several recent studies \cite{mahadevan2023cost,vzliobaite2015towards,regol2025when} have explored the trade-offs between model staleness and retraining costs, they primarily target drift detection or updating strategies tailored to specific models or domains. In all, existing approaches rarely consider how the statistical properties of drift events influence elastic compute allocation and deployment scheduling -- gaps that our work addresses through a general, model-agnostic framework.

Finally, our problem lies within the scope of MLOps -- the practice of reliably developing, deploying, and maintaining machine learning systems at scale  \cite{kreuzberger2022machinelearningoperationsmlops,muravevMLOpsArchitectureFuture,stone2025navigating}. While MLOps primarily aims to establish best practices for ML system operation, the growing complexity and escalating costs of modern ML pipelines \cite{cottier2025risingcoststrainingfrontier} have spurred interest in cost-aware and compute-efficient retraining and deployment strategies \cite{yu2022gadget,madireddy2019adaptive,shen2024cost,wang2021elastic,qiao2021pollux}. These developments further underscore the need for formal frameworks that link resource allocation to model performance under dynamic data conditions -- the focus of our study.

\subsection{Contributions}

Motivated by the need to sustain performance in distributed ML systems operating under resource constraints and non-stationary data conditions, we develop a model-agnostic framework that captures the interplay between elastic resource allocation and sudden concept drift. Our main contributions are:

\begin{enumerate}

\item We propose a model-agnostic analytical framework that characterizes the performance of both client-side and server-side models, capturing the effects of training resource allocation, sudden concept drift, and deployment policy. The proposed framework enables formal analysis of how these factors jointly influence model performance.

\item We show that optimal training resource allocation policies—those that maximize the long-term performance of server-side models—are of the bang-bang control type, i.e., they switch between extreme values. We prove that a single-switch front-loading policy is optimal if and only if concept durations follow a Decreasing Mean Residual Life (DMRL) distribution. We further demonstrate that intuitive allocation heuristics are sub-optimal under Increasing Mean Residual Life (IMRL) concept duration distributions.

% We derive optimal training resource allocation policies for concept duration following Decreasing Mean Residual Life (DMRL) distribution, aiming to maximize the long-term performance of server-side models. We further demonstrate the sub-optimality of intuitive allocation heuristics under Increasing Mean Residual Life (IMRL) concept durations.

\item We analyze the deployment scheduling problem under communication constraints and establish quasi-convexity of the performance objective under mild conditions. We derive necessary optimality conditions and introduce a randomized deployment policy that achieves target deployment rates with theoretical performance guarantees.

\end{enumerate}

Together, these contributions establish theoretical foundations and algorithmic principles for designing resource-aware training and deployment policies in large-scale, real-world ML systems.

The remainder of the paper is organized as follows. Section 2 formalizes the system model, including concept drift dynamics and performance metrics. Section 3 analyzes optimal training resource allocation under budget constraints and characterizes how different aging properties of concept durations affect the resulting policies. Section 4 turns to deployment scheduling under communication constraints and develops near-optimal, drift-aware strategies. Section 5 offers concluding remarks.

%\section{Framework for Resource Allocation for Training and Deployment under Sudden Concept Drift}

\section{A Framework for Resource Allocation under Sudden Concept Drift}
\label{sec:system_model}

%We study distributed machine learning (ML) systems under sudden concept drift and introduce a mathematical framework for analyzing how training and deployment resource allocation impact systems performance.

We consider a system that consists of a central server, referred to as the ML service provider, that possesses substantial computational resources and access to large-scale datasets, and multiple client devices with limited local capabilities. While clients can perform inference using deployed models, they cannot retrain or update them independently. The provider is therefore responsible for monitoring model performance, retraining as necessary, and redeploying updated models in response to distributional shifts that degrade accuracy on both the server and client sides. In this work, we focus specifically on modeling and managing the effects of \textit{sudden concept drift}, where data distributions shift abruptly, causing immediate drops in model performance.

A model’s performance depends on several factors, including its complexity, the quality and quantity of training data, and the computational resources allocated to training. As training progresses and model parameters are updated, performance typically improves. However, evaluation metrics such as accuracy, F1-score, or loss fluctuate due to the stochastic nature of optimization methods (e.g., stochastic gradient descent and its variants) and the limited size of evaluation datasets. Consequently, these metrics offer only an approximate view of the model's true performance on the underlying data distribution. Under concept drift, where the distribution itself evolves, performance assessment becomes even more challenging, motivating the need for more robust and interpretable measures of model quality.

To address these challenges, our framework introduces an idealized performance metric, the \textit{expected concept loss}, defined as the model’s expected loss over the true data distribution corresponding to the current concept. This loss is assumed to decrease monotonically during training, with the rate and shape of improvement determined by the model architecture, optimization algorithm, data characteristics, and allocated resources. Under sudden concept drift, the trajectory of expected concept loss changes abruptly between concepts as the underlying distribution shifts. Let \(T_i\) denote the start time of the \(i^{\text{th}}\) concept, and let \(Y_i\) represent its random duration, such that \(T_{i+1} - T_i = Y_i\) with $\E[Y]<\infty$. We model the evolution of the expected concept loss during concept $i$ by a random function \(G_i(\cdot): \mathbb{R}^+ \rightarrow \mathbb{R}^+\), drawn from a countable set \(\mathcal{G}\) of convex, decreasing functions. Here, \(G_i(0)\) denotes the loss immediately following the onset of concept \(i\), and \(G_i(t)\) describes the decay in loss over time under unit resource allocation. The overall concept dynamics are captured by the stochastic process \(((Y_i, G_i(\cdot)) : i \in \mathbb{N})\), where the pairs \((Y_i, G_i(\cdot))\) are assumed to be independent and identically distributed (i.i.d). Although this formulation primarily targets sudden drift, it also accommodates mixed settings where sudden and gradual drifts coexist. In practical implementations, a change detection algorithm would identify points of significant distributional shift, triggering retraining when a change is detected. Our model abstracts this mechanism by capturing the timing and magnitude of such discrete or sufficiently large gradual changes that warrant retraining.

The amount of training resources determines both how quickly a model learns and how well it ultimately performs. In supervised learning over a fixed dataset, greater compute capacity enables faster model parameter updates. In online learning, the training rate is limited by the data arrival process, which can sometimes be accelerated, e.g., by increasing sensor frequency or acquiring additional labels through crowdsourcing \cite{huang2025context,cheng2018learning,mandlekar2018roboturk,bang2022online}. Our framework unifies these settings under the common notion of \textit{training resource allocation}.

We assume that training resources can be adjusted dynamically over time. Let \(e_i(t): [0, y_i) \rightarrow [0, M]\) denote the resource allocation during concept \(i\), where $M$ is the maximum allowable resource level and $y_i$ is a realization of the concept duration $Y_i$. Training speed is assumed to scale linearly with $e_i(t)$, which we model through a horizontal scaling transformation of \(G_i(\cdot)\) controlled by the allocation function $e_i(\cdot)$. The resulting server-side expected concept loss at time $t$, denoted $\mathcal{L}(t)$, reflects both the stochastic concept drift process $(Y_i, G_i(\cdot))$ and the corresponding resource allocation $e_i(\cdot)$, and is formalized as

\begin{equation}
\mathcal{L}(t)
= G_{I(t)}\left(\int_{0}^{t-T_{I(t)}} e_{I(t)}(\tau) d\tau\right),
\label{eq:modified_expected_loss}
\end{equation}
where $I(t) = \max \{i \in \mathbb{N} : T_i \le t\}$ denotes the index of the concept active at time $t$. Since both $G_{I(t)}$ and $I(t)$ are random, $\mathcal{L}(t)$ is itself a stochastic process. The quantity \(\mathcal{L}(t)\) captures the expected concept loss of the server-side model under continuous training. In contrast, client devices operate on previously deployed versions that are updated only at discrete deployment times. Consequently, the client-side expected loss, denoted $\hat{\mathcal{L}}(t)$, generally lags behind and may differ from $\mathcal{L}(t)$.

Let $D_{i,j}$ denote the time of the $j^{\text{th}}$ model deployment during concept $i$, with the first deployment occurring at the start of the concept, $D_{i,0} = T_i$. At any time $t$, client devices operate the model version most recently deployed, i.e., the one corresponding to the latest deployment at or before $t$. The expected concept loss experienced by clients at time $t$, denoted $\hat{\mathcal{L}}(t)$, is therefore given by
\begin{equation}
\hat{\mathcal{L}}(t) = G_{I(t)}\left(\int_{0}^{D_{I(t),J(t)}-T_{I(t)}} e_{I(t)}(\tau) d\tau\right),
\label{eq:deployed_expected_loss}
\end{equation}
where $D_{I(t),J(t)} = \max \{D_{i,j} : D_{i,j} \leq t\}$ is the most recent deployment time prior to $t$.

\begin{figure}
   \centering
   \includegraphics[width=0.8\linewidth]{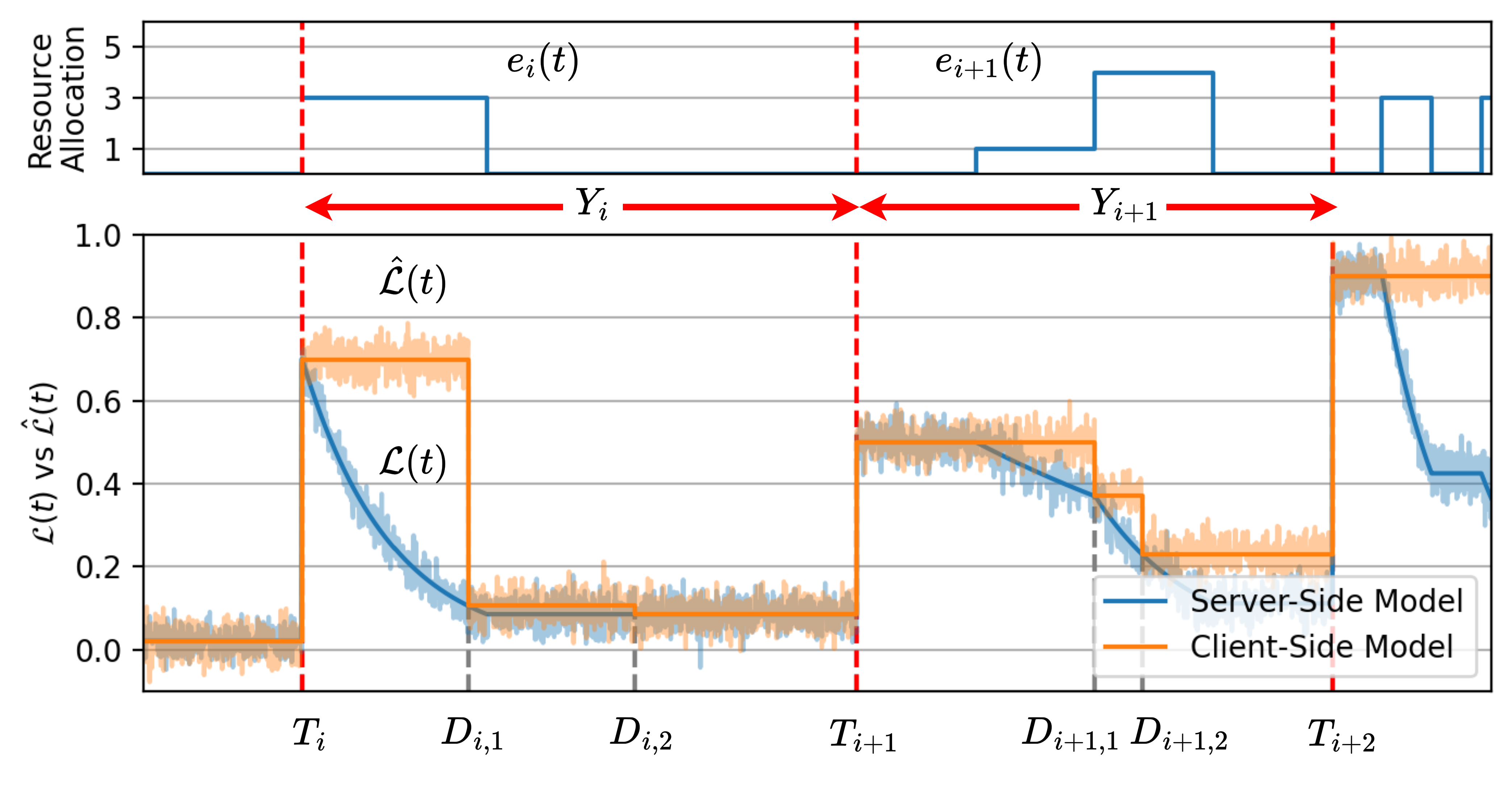}
   \caption{Top: training resource allocation $e_i(t)$. Bottom: sample paths of the expected concept loss for the server $\mathcal{L}(t)$ and client models $\hat{\mathcal{L}}(t)$, as defined in \eqref{eq:modified_expected_loss} and \eqref{eq:deployed_expected_loss}. Shaded regions indicate fluctuations in observed evaluation loss. $T_i$ denotes concept change times, and $D_{i,j}$ denote deployment times.}
    \label{fig:sample_path}
\end{figure}

% \begin{wrapfigure}{r}{0.6\textwidth}
%   \centering
%   \includegraphics[width=0.6\textwidth]{figures/samplepath_variation.drawio-3.png}
%   \caption{Top: allocated training resources $e_i(t)$. Bottom: sample paths of expected concept loss at the server $\mathcal{L}(t)$ and clients $\hat{\mathcal{L}}(t)$, as defined in \eqref{eq:modified_expected_loss} and \eqref{eq:deployed_expected_loss}.}
%   \label{fig:sample_path}
% \end{wrapfigure}

Figure \ref{fig:sample_path} illustrates the evolution of the expected loss under a given training resource allocation. The top plot shows the allocated resources $e_i(t)$, which determine the training speed. In the bottom plot, the blue curve depicts the expected concept loss at the server, $\mathcal{L}(t)$. It rises sharply at each concept change and then decays -- at a constant rate during concept $i$ and at a varying rate during concept $i+1$ -- reflecting the change in resource allocation. The orange curve shows the expected loss of the deployed client-side model, which is updated to match the server-side model at deployment times. The fluctuations around both curves represent variations in observed performance metrics computed on sampled evaluation data.

The above framework captures how training and deployment resource allocations influence both server-side and client-side performance under sudden concept drift.
The key assumptions are summarized as follows:
\begin{itemize}
    \item The expected concept loss remains constant during periods when no training resources are allocated.
    \item Upon a concept change, data from the new distribution becomes immediately available, allowing training to begin without delay if resources are allocated.
    \item Each concept persists for an i.i.d random duration $Y_i$ with bounded expectation, $\E[Y] < \infty$.
    \item The expected loss decay for concept $i$ is modeled by a decreasing convex function $g_i(\cdot): \mathbb{R}^+ \rightarrow \mathbb{R}^+$, which is a realization of $G_i(\cdot)$.
    \item Training resources are continuously adjustable over time.
    \item Allocated resources $e_i(t) \in [0,M]$ scale the training speed linearly.
    \item At each concept change, the expected concept loss of both the deployed and server-side models resets to a common initial level.
\end{itemize}
% (a) Expected concept remains constant in periods where no training resources are allocated. (b) Upon a concept changes, data from the new distribution becomes immediately available, enabling training to begin without delay if resources are allocated. (c) Each concept lasts for an i.i.d random duration $Y_i$ with bounded expectation, $\E[Y] < \infty$.
% (d) The expected loss decay for concept $i$ is modeled by a decreasing convex function $g_i(\cdot): \mathbb{R}^+ \rightarrow \mathbb{R}^+$, a realization of $G_i(\cdot)$.
% (e) Training resources are continuously adjustable over time. (f) Allocated resources $e_i(t) \in [0,M]$ scale training speed linearly. (g) At each concept change, the expected concept loss of both the deployed and server models resets to the same initial level.

Next, we investigate two key questions: (1) Training resource allocation: Given a constraint on total training resources, how should they be allocated over time to mitigate the effects of sudden concept drift and optimize server-side model performance? (Section~\ref{sec:training}) (2) Deployment scheduling: Under a constraint on deployment frequency, what constitutes the optimal deployment policy that maximizes client-side performance under concept drift? (Section~\ref{sec:deployment})

% The next section formalizes the training resource allocation problem and derives optimal allocation strategies under sudden concept drift.

%%%%%%%%%%%%%%%%%%%%%%%%%%%%%%%%%%%%%%%%%
\section{Optimizing Resource Allocation for Training under Concept Drift}
%\section{Training Resource Allocation Problem for Minimizing Time-Average expected concept loss}
% XXGDV Original size 3.5 pages.
\label{sec:training}

In this section, we study resource allocation policies that optimize the time-average performance of the ML model on the server side, subject to a time-average budget constraint and in the presence of sudden concept drift, as defined in the framework of Section \ref{sec:system_model}. We turn to model deployment and the resulting client-side performance in Section \ref{sec:deployment}.

{\bf Compute budget constraint on training resources.}

As noted earlier, we assume that the ML service provider has access to elastic compute resources, i.e., it can dynamically purchase training capacity over time. We adopt a pay-as-you-go pricing model, where resources are billed at a fixed rate $\sigma_e$ per unit of time and resource, and the provider is subject to a fixed average cost budget constraint denoted by $B$. Specifically, given the resource allocation functions $e_i(t):[0,\infty) \rightarrow [0,M]$ over the duration of each concept $i$, the time-average budget constraint is expressed as
\begin{equation}
\label{eq:time-average_constraint}
    \limsup_{t \rightarrow \infty} \frac{\sigma_e}{t} \int_{\tau=0}^t e_{I(\tau)}(\tau-T_{I(\tau)}) d\tau  \leq B,
\end{equation}
where $I(\tau)$ denotes the index of the current concept at time $\tau$, with arrival time $T_{I(\tau)}$.

%Subject to the cost budget constraint \eqref{eq:time-average_constraint},
{\bf Server-side time-average expected loss.} Our objective is to determine a resource allocation policy that minimizes the time-average expected loss of the server-side model, given by
\begin{align}
    \limsup_{t \rightarrow \infty} \frac{1}{t} \int_{\tau=0}^t \mathcal{L}(\tau) d\tau&=
   % \limsup_{t \rightarrow \infty} \frac{1}{t} \int_{\tau=0}^t G_{I(\tau)}\left(\int_{u=0}^{\tau-T_{I(\tau)}} e_{I(\tau)}(u) du\right) d\tau \\
   % &=
   \limsup_{t \rightarrow \infty} \frac{1}{t} \left[ \sum_{i=1}^{I(t)-1} \int_{\tau = 0}^{Y_i} G_i(\int_0^\tau e_i(u)du) d\tau \right] \label{eq:time-average_objective}
\end{align}
where $G_i(\cdot)$ denotes the random expected concept loss function associated with concept $i$, and $Y_i = T_{i+1} - T_i$ is the duration of that concept. The equality follows by decomposing the time-average loss into a sequence of cycles corresponding to concept durations and their associated losses.

As evident from the objective function in \eqref{eq:time-average_objective}, the optimal training resource allocation $e_i(t)$ may depend on both the expected loss function $g_i(\cdot)$ and the duration $y_i$ of concept $i$. Determining such an optimal policy would generally require accurate knowledge of both $g_i(\cdot)$ and $y_i$, which is often infeasible in real-world settings. To develop a more practical and analytically tractable formulation, we consider stationary {\em static training resource allocation policies}, wherein the same allocation function is applied across all concept periods, independent of concept-specific characteristics $g_i(\cdot)$ and $y_i$.

% In this paper we shall consider stationary {\em static training resource allocation policies}, where the same resource allocation function is applied across all concept periods, independent of their concept-specific characteristics, $g_i(\cdot)$ and $y_i$.

\begin{definition}[Static training resource allocation policy]
A collection of resource allocation functions $\{e_i(\cdot): i \in \mathbb{N}\}$ specifies a training resource allocation policy.
Such a policy is said to be \emph{static} -- and therefore stationary and causal -- if, for all concepts $i \in \mathbb{N}$ and all $t \in [0, \infty)$, it holds that $e_i(t) = e(t)$.
\end{definition}

We assume a static training resource allocation policy \(e(\cdot)\), and that the pairs \(((Y_i, G_i(\cdot)) : i \in \mathbb{N})\), where each \(Y_i\) denotes the concept duration and \(G_i(\cdot)\) the corresponding expected concept loss function, are independent and identically distributed. Under these assumptions, the limits in \eqref{eq:time-average_objective} and \eqref{eq:time-average_constraint} exist. By the Renewal-reward Theorem \cite[Section 3.4]{gallagerDiscreteStochasticProcesses1996}, the server-side time-average expected loss is given by
\begin{align}
    \limsup_{t \rightarrow \infty} \frac{1}{t} \int_{\tau=0}^t \mathcal{L}(\tau) d\tau &=
    %\frac{\E [ \int_{\tau = 0}^{Y_i} G_i(\int_0^\tau e(u)du) d\tau]}{\E [Y_i]} \label{eq:renewal_GY_0}= \frac{\E [ \int_{\tau = 0}^{Y_i} \E [G_i(\int_0^\tau e(u)du)| Y_i] d\tau]}{\E [Y_i]} \\
    %&=
    \frac{\E [ \int_{\tau = 0}^{Y} \bar{g}(\int_0^\tau e(u)du) d\tau]}{\E [Y]} \label{eq:renewal_GY},
\end{align}
where the equality follows from the independence of $G_i$ and $Y_i$\footnote{Although the time-averages in \eqref{eq:renewal_GY} are derived under the assumption that $G_i$ and $Y_i$ are independent, the result extends to more general cases in which the limit exists with a continuous, convex, decreasing function $\bar{g}: [0,\infty) \rightarrow \mathbb{R}^+$, not necessarily equal to the mean $\E[G_i(\cdot)]$.}. We define $\bar{g}(\cdot) = \E[G_i(\cdot)|Y_i]=\E[G_i(\cdot)]$ as the expected loss function, representing the average expected concept loss across different concepts. Since each $g_i(\cdot)$ is assumed convex and decreasing, their expectation $\bar{g}(\cdot)$ inherits these properties\footnote{The convexity assumption on individual expected concept loss functions $g_i(\cdot)$ can be relaxed, since our analysis only requires the convexity of the aggregated expected loss curve $\bar{g}(\cdot)$ under static training resource allocation.}. The corresponding time-average budget constraint then becomes
\begin{equation}
    \label{eq:time-average_constraint_computed}
    \limsup_{t \rightarrow \infty} \frac{\sigma_e}{t} \int_{\tau=0}^t e_{I(\tau)}(\tau-T_{I(\tau)}) d\tau   = \sigma_e \frac{\E [\int_0^Y e(t) dt]}{\E [Y]} \leq B.
\end{equation}

To determine the optimal static resource allocation policy that minimizes the time-average expected loss in \eqref{eq:renewal_GY} while satisfying the budget constraint in \eqref{eq:time-average_constraint_computed}, we rewrite both expressions using Fubini’s Theorem and obtain the infinite-horizon continuous-time optimal control problem
\begin{align}
\min_{e(\cdot)} \quad & J(e) = \int_0^\infty \bar{g}(x(t)) \bar{F}_Y(t)dt \label{opt:reformulated_problem}\\
\textrm{s.t.} \quad & x'(t) = e(t), \quad x(0)=0, \quad 0 \leq e(t) \leq M, \quad \forall t \in [0,\infty)\\
& C(e) = \int_0^\infty e(t)\bar{F}_Y(t)dt \leq B_1, \quad \text{where } B_1 = \frac{B \cdot E[Y]}{\sigma_e}. \label{eq:reformulated_constraint1}
\end{align}
Here, $\bar{F}_Y(t) = \mathbb{P}(Y > t)$ denotes the survival function (i.e., the complementary cumulative distribution function) of the concept duration $Y$.
The optimization is taken over static resource allocation functions $e(t)$ that are measurable and bounded on $[0, \infty)$, lying in the admissible set
\(\mathcal{E} = \{ e : e(t) \in [0, M],\ \forall t \ge 0 \}\).
For any $e \in \mathcal{E}$, we define the state variable $x(t) = \int_{0}^{t} e(\tau)\, d\tau$, which is absolutely continuous.

%Here, $\bar{F}_Y(t) = \mathbb{P}(Y > t)$ denotes the survival function (i.e., the complementary cumulative distribution function) of the concept duration $Y$. The constant $\mathbb{E}[Y]$ in the denominator of \eqref{eq:time-average_objective} is omitted from the optimization since it does not affect the solution.

\begin{lemma}[Existence of an optimal solution for Problem~\eqref{opt:reformulated_problem}]
\label{lem:convexity}
Let $\bar{g}: [0,\infty) \to \mathbb{R}^+$ be continuously differentiable, convex, and non-increasing.
Then the functional optimization problem \eqref{opt:reformulated_problem} is convex and admits at least one optimal solution.
In particular, the functional $J(e)$ defined in \eqref{opt:reformulated_problem} and the feasible set of resource allocations
\[
\mathcal{E}_C = \left\{ e \in \mathcal{E} \;\middle|\; \int_{0}^{\infty} e(t)\, \bar{F}_Y(t)\, dt \le B_1 \right\}
\]
are both convex over the domain
\(\mathcal{E} = \{ e : e(\tau) \in [0, M],\ \forall \tau \ge 0 \}\).
\end{lemma}
The proof of Lemma~\ref{lem:convexity} follows from the linearity of the state function
\(x(t) = \int_{0}^{t} e(\tau)\, d\tau\), the convexity of \(\bar{g}(x)\), and the non-negativity of the survival function \(\bar{F}_Y(t)\).
Complete proof details are provided in Appendix~\ref{proof:lem:convexity}.

% Further as discussed in more detail in  Appendix \ref{supp:PMP_conditions} it follows by Arrow’s Sufficiency Theorem \cite[Theorem 3.1]{sethiOptimalControlTheory2021}, that the necessary conditions provided by Pontryagin’s Maximum Principle (PMP) are also sufficient for global optimality in this setting.
% This in turn allows us to conclude imply that the optimal  policy takes the form of a "bang-bang" control, i.e., transitions from one extreme control value to another.

Lemma~\ref{lem:convexity} establishes the existence of an optimal solution to Problem~\eqref{opt:reformulated_problem}. Moreover, by Arrow’s sufficiency theorem~\cite[Theorem~3.1]{sethiOptimalControlTheory2021}, the necessary conditions provided by Pontryagin’s Maximum Principle (PMP) are also sufficient for global optimality in this setting.

\subsection{Application of Pontryagin's Maximum Principle to Problem~\ref{opt:reformulated_problem}}

To analyze the structure of the optimal solution to the problem in \eqref{opt:reformulated_problem}, we apply Pontryagin’s Maximum Principle (PMP).
We define the Hamiltonian function \(H(t, x, e, p, \nu)\), which combines the integrand from the objective \eqref{opt:reformulated_problem}, a constant Lagrange multiplier \(\nu \ge 0\) associated with the isoperimetric constraint \eqref{eq:reformulated_constraint1}, and the costate variable \(p(t)\) corresponding to the state dynamics as
\begin{equation}
    H(t, x, e, p, \nu) = \bar{g}(x(t))\bar{F}_Y(t) + (p(t) + \nu \bar{F}_Y(t))e(t).
\end{equation}
The necessary and sufficient conditions (by Lemma~\ref{lem:convexity}) given by Pontryagin’s Maximum Principle (PMP) for the infinite-horizon optimal control problem with free terminal state are as follows~\cite[Sec.~3.6]{sethiOptimalControlTheory2021}:
\begin{align}
x'(t) &=  \frac{\partial H}{\partial p} = e(t), \quad x(0)=0 \\
p'(t) &= -\frac{\partial H}{\partial x} = -\bar{g}'(x(t))\bar{F}_Y(t) \label{eq:costate_eqn_main} \\
e^*(t) &= \argmin_{e \in \mathcal{E}} H(t, x, e, p, \nu) = \argmin_{e \in \mathcal{E}} (p(t) + \nu \bar{F}_Y(t))e(t) \label{eq:control_opt_main} \\
\lim_{t\to\infty} p(t) &= 0 \label{eq:transversality_main} \\
\nu \ge 0, \quad &\nu \left( \int_0^\infty e^*(t)\bar{F}_Y(t)dt - B_1 \right) = 0. \label{eq:complementary_slackness_main}
\end{align}
The transversality condition~\eqref{eq:transversality_main} follows from the boundedness of the resource allocation function, \(e(t) \in [0,M]\), and the finiteness of the expected concept duration, \(\mathbb{E}[Y] < \infty\).
Since the Hamiltonian is linear in the control variable \(e(t)\), the optimal resource allocation policy must take the form of a \emph{bang-bang} or singular control.
We define the corresponding switching function as
\(\phi(t) \coloneqq p(t) + \nu\, \bar{F}_Y(t)\). From~\eqref{eq:control_opt_main}, the optimal resource allocation policy is given by
\begin{equation} \label{eq:control_law_main_final}
e^*(t) = \begin{cases} 0 & \text{if } \phi(t) > 0 \\ M & \text{if } \phi(t) < 0 \\ e_s(t) \in [0,M] & \text{if } \phi(t) = 0 \text{ over an interval (singular control).} \end{cases}
\end{equation}
By the transversality condition~\eqref{eq:transversality_main} and integration of the costate equation~\eqref{eq:costate_eqn_main}, we obtain $p(t) = \int_t^{\infty} \bar{g}'(x(s))\bar{F}_Y(s)ds$. Substituting this expression into the switching function yields
\begin{equation}
    \phi(t) = \int_{t}^{\infty} \bar{g}'(x(s))\bar{F}_Y(s)ds + \nu \bar{F}_Y(t).
\end{equation}
Furthermore, using~\eqref{eq:costate_eqn_main} and the identity
$\bar{F}'_Y(t) = - h_Y(t)\bar{F}_Y(t) $, the derivative of the switching function becomes
\begin{equation}
    \phi'(t) = p'(t) + \nu \bar{F}'_Y(t)= \bar{F}_Y(t) [-\bar{g}'(x(t)) - \nu h_Y(t)]. \label{eq:derivative_switching}
\end{equation}
These expressions form the basis for analyzing the monotonicity of the switching function.
% In many control problems with monotonic dynamics, the optimal policy often exhibits a single switching time, transitioning from one extreme control value to another. In our setting, this behavior arises under specific structural assumptions about the concept duration distribution $Y$, particularly when it exhibits monotonic aging characteristics such as Decreasing Mean Residual Life (DMRL) or Increasing Mean Residual Life (IMRL).

As we shall see, the characteristics of the concept duration distribution determine whether the optimal policy exhibits a single switching time.
We discuss these characteristics below.

\begin{definition}[Mean Residual Life (MRL)] The \emph{mean residual life} (MRL) function $m_Y(t)$ of a nonnegative random variable $Y$ is defined as the expected remaining lifetime given survival up to time $t$:
\begin{equation}
     m_Y(t) \coloneqq E[Y-t | Y > t] =  \frac{\int_t^\infty \bar{F}_Y(u)du}{\bar{F}_Y(t)}, \quad \text{for } \bar{F}_Y(t)>0.
\end{equation}
A well-known relationship (see~\cite{barlow1974}) links the derivative of the MRL function and the hazard rate $m'_Y(t) = h_Y(t)m_Y(t) - 1$, where $h_Y(t) \coloneqq \frac{f_Y(t)}{\bar{F}_Y(t)}$, for $\bar{F}_Y(t)>0$. Accordingly, the random variable $Y$ is said to have
\begin{description}
    \item[]Decreasing Mean Residual Life (DMRL) if $m_Y(t)$ is non-increasing, i.e., $h_Y(t)m_Y(t) \leq 1$.
    \item []Increasing Mean Residual Life (IMRL) if $m_Y(t)$ is non-decreasing, i.e., $h_Y(t)m_Y(t) \geq 1$.
\end{description}
\end{definition}
Many real-world processes naturally exhibit these aging characteristics.
DMRL behavior is commonly observed in systems that degrade over time, such as mechanical components, electronic devices, or biological organisms, where aging applies intuitively.
In contrast, IMRL behavior characterizes processes in which survival implies increased robustness or reliability, such as the lifecycles of businesses or startups, social and political trends, and the adoption of new technologies or products.
Accordingly, depending on the application, the aging properties of concept durations can often be assumed or empirically verified using sample-efficient statistical methods~\cite{bergmanFamilyTestStatistics1989, Hall2020, kocharEstimationMonotoneMean2000}.

% If concept durations have the DMRL property we shall the optimal resource allocation policy is a {\em Front-loading policy}, i.e., allocates a maximum amount of resources at the start of each concept duration up to some fixed time--formally stated in the following theorem shown in the Appendix \ref{proof:back-loading}.

Next, we analyze the optimality of single-switch policies given the MRL properties of the concept duration distribution. A single-switch resource allocation policy can take one of two forms: (i) \textit{Front-Loading} policy, where $e(t) = M$ for $t \in [0,t^*]$ and $e(t)=0$ for $t > t^*$; or (ii) \textit{Back-Loading} policy, where $e(t) = 0$ for $t \in [0,t^*]$ and $e(t)=M$ for $t > t^*$. Here, $t^*$ denotes the switching time. For such policies to be optimal, the switch at time $t^*$ must satisfy $\phi(t^*) = 0$, and the sign of the switching function should be opposite on the intervals before and after the switch, while remaining constant within each interval. The following theorems characterize the conditions under which the front-loading or back-loading policy is optimal, depending on the aging characteristics (DMRL or IMRL) of the concept duration distribution.

\begin{theorem}[Optimality of the Front-Loading Policy]
\label{theorem:front-loading}
For any convex, decreasing expected loss function $\bar{g}: [0,\infty) \rightarrow \mathbb{R}^+$ and any resource budget $B > 0$,
the optimal resource allocation is a single-switch control of the form
\begin{equation}
    e^*(t) = \begin{cases} M & t<t_{DMRL}^* \\
0 & t \geq t_{DMRL}^*,
\end{cases}
\label{eq:opt_res_alloc}
\end{equation}
if and only if the concept duration $Y$ has a decreasing mean residual life (DMRL). The switching time $t^*_{DMRL}$ is uniquely determined by the budget constraint $\int_0^{t_{DMRL}^*} \bar{F}_Y(y)dy = \frac{\mathbb{E}[Y] B}{ M \sigma_e}$, provided that $B<M\sigma_e $, and $t^*_{DMRL} = \infty$, otherwise.
% $$\int_0^{t_{DMRL}^*} \bar{F}_Y(y)dy = \frac{\mathbb{E}[Y] B}{ M \sigma_e},$$
% and $t^*_{DMRL} = \infty$, otherwise.
\end{theorem}
The proof of Theorem~\ref{theorem:front-loading} relies on analyzing the behavior of the switching function $\phi(t)$ under DMRL concept duration distributions.
By leveraging the convexity of the expected loss function $\bar{g}(\cdot)$ and the monotonic aging property of $Y$, we show that $\phi(t)$ crosses zero at most once, and that its sign pattern aligns with the optimality structure specified in~\eqref{eq:control_law_main_final}.
The complete proof is provided in Appendix~\ref{proof:front-loading}.

Note that Theorem~\ref{theorem:front-loading} states that if $Y$ is DMRL, the optimal policy is front-loading, and conversely, if the optimal policy is front-loading for all budgets, then $Y$ must be DMRL.
While such policies may appear intuitive, this result formally rules out the optimality of commonly used heuristics such as fixed or periodic training schedules.
Moreover, it implies that if $Y$ is not DMRL, a front-loading policy need not be optimal for all budget levels.
The next result, proved in Appendix~\ref{proof:idling}, shows that when $Y$ is IMRL, the optimal policy exhibits delayed (i.e., idling) resource allocation at the start of each concept duration.

\begin{theorem}[Optimality of the Back-Loading Policy]
\label{theorem:idling}
For any convex, decreasing expected loss function $\bar{g}: [0,\infty) \rightarrow \mathbb{R}^+$,
if the concept duration $Y$ has an increasing mean residual life (IMRL) and the budget satisfies $B < M\sigma_e$,
the optimal resource allocation exhibits an initial idling period before training begins,
\begin{equation}
e^*(t) = 0, \quad \text{for } t < t^*_{IMRL},
\end{equation}
where $t^*_{IMRL} > 0$ denotes the time at which resource allocation starts.
\end{theorem}

%Similar to Theorem \ref{theorem:front-loading}, the proof of Theorem \ref{theorem:idling} leverages the optimality of the bang-bang control for Problem \eqref{opt:reformulated_problem} and the monotonic aging property of concept duration distribution. A complete proof is provided in Appendix \ref{proof:idling}.

% Additionally, if the expected loss curve is decreasing and linear one can show that the optimal policy Back-loading, i.e, idles and then switches to allocating a maximum. See Corollary \ref{corollary:back-loading} included and proven in  Appendix \ref{proof:back-loading}.

Although the IMRL case may represent an edge scenario, Theorem~\ref{theorem:idling} characterizes the conditions under which the back-loading policy is optimal. The following corollary provides sufficient conditions for the optimality of such a resource allocation policy.
\begin{corollary}[Optimality of the Back-Loading Policy]
\label{corollary:back-loading}
For any convex, decreasing expected loss function $\bar{g}:[0,\infty]\rightarrow \mathbb{R}^+$ with constant derivative $\bar{g}'(t) = -\beta$, and any concept duration $Y$ with an increasing mean residual life (IMRL), the optimal resource allocation is a single-switch control given by
\begin{equation}
   e^*(t) = \begin{cases} 0 & t \leq t_{IMRL}^* \\
M & t > t_{IMRL}^*
\end{cases}
\label{eq:opt_res_alloc}
\end{equation}
where the switching time $t_{IMRL}^*$ is uniquely determined by the budget constraint
$$\int_{t_{IMRL}^*}^{\infty} \bar{F}_Y(y)dy = \frac{\mathbb{E}[Y] B}{ M \sigma_e}, $$
provided that $B<M\sigma_e$, and $t^*_{IMRL} = 0$ otherwise.
\end{corollary}
The proof of Corollary \ref{corollary:back-loading} follows directly from Theorem \ref{theorem:idling} and the fact that a linearly decreasing $\bar{g}(\cdot)$ ensures that the switching function crosses zero at most once. The complete proof is provided in Appendix \ref{proof:back-loading}.

Notably, the strength of Theorems \ref{theorem:front-loading} and \ref{theorem:idling} lies in their generality: their conclusions do not depend on the specific functional form of $\bar{g}(\cdot)$, but only on its monotonicity and convexity. These results also demonstrate that training policies that ignore the aging behavior of concept durations are provably suboptimal. The DMRL/IMRL distinction further clarifies how drift dynamics shape the structure of optimal resource allocation policies. Even without precise distributional information, our analysis indicates when front-loading (DMRL) or deferral (IMRL) strategies are most effective. As such, these results provide broadly applicable guidance for designing training resource allocation policies based on the statistical characteristics of training cycles driven by sudden concept changes.

\begin{figure}[h]
    \centering
    \begin{subfigure}{0.48\linewidth}
        \centering
        \includegraphics[height=4.5cm]{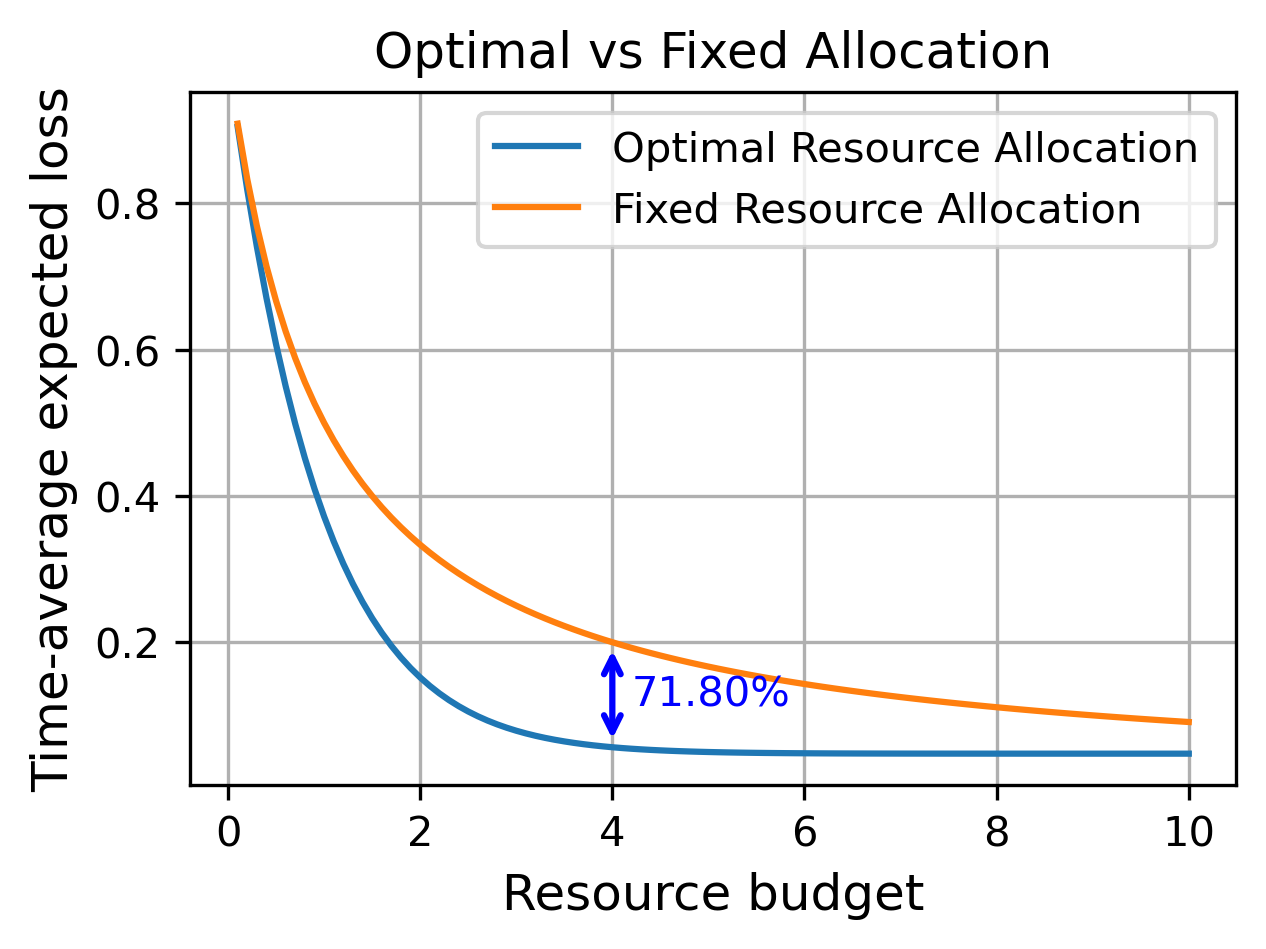}
        \caption{
        Comparison between constraint-binding optimal and fixed resource allocation policies.
        $Y \sim \mathrm{Exp}(1)$, $\bar{g}(t)=e^{-t}$, $\sigma_e=1$, $M=20$.
        %The optimal (front-loading) policy achieves up to $71.8\%$ lower time-average expected loss.
        }
        \label{fig:opt_vs_fixed}
    \end{subfigure}
    \hfill
    \begin{subfigure}{0.48\linewidth}
        \centering
        \includegraphics[height=4.5cm]{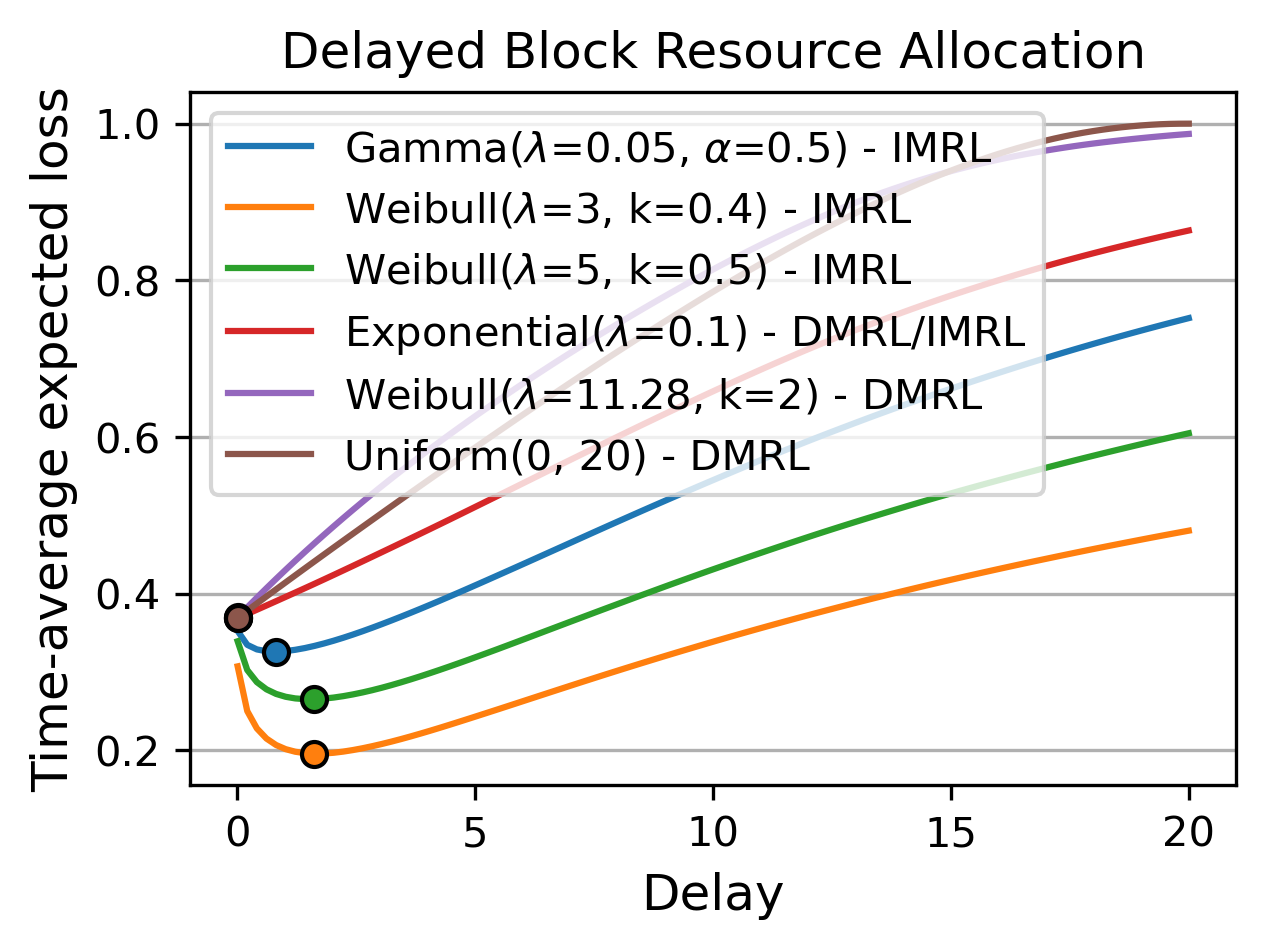}
        \caption{
        Constraint-binding delayed block allocation with varying delays under different concept duration distributions.
        $\bar{g}(t)=e^{-t}$, $B=0.1$, $\sigma_e=1$, $M=20$.
        %IMRL distributions (e.g., Gamma, Weibull with $k<1$) benefit from delay, whereas DMRL distributions degrade.
        }
        \label{fig:delayed}
    \end{subfigure}

    \caption{
    Comparison of optimal, fixed and delayed resource allocation policies under sudden concept drift.
    All simulations were executed on a MacBook Air~M3 with 16\,GB of RAM.
    }
\end{figure}

%%%%%%%%

In Figures \ref{fig:opt_vs_fixed} and \ref{fig:delayed}, we present simulation results illustrating the benefits of optimal training resource allocation and validating the relationship between the aging properties of concept durations and the structure of optimal allocation policies. In Figure \ref{fig:opt_vs_fixed}, we compare the time-average expected loss under two policies: the optimal resource allocation (specifically, the front-loading policy derived from Theorem \ref{proof:front-loading}) and a fixed resource allocation. In this experiment, concept durations follow an exponential distribution, corresponding to the DMRL case, and the expected loss function decays exponentially. The fixed policy assigns a constant resource level such that the budget constraint is exactly met, i.e., $e_{\text{fixed}}(t) = B/\sigma_e$ $\forall t$. The results show that, for the same resource budget, the optimal policy reduces the time-average expected loss by up to $71.80\%$ compared to the fixed allocation policy. As the budget $B$ approaches the upper limit $M\sigma_e$, the switching time $t^*_{\text{DMRL}}$ increases and the performance gap between the two policies narrows. Additional simulation results using alternative expected loss functions are provided in Appendix \ref{app:additional_simulations}.

In Figure \ref{fig:delayed}, we examine the effect of delayed block resource allocation for concept duration distributions with different aging properties. These experiments also use an exponentially decaying expected loss function, and the parameters are chosen such that the budget constraint is binding. Under a delayed block allocation policy, resources are withheld until a delay $z$, after which the full capacity is applied until the budget is depleted. Formally, the allocation is defined as
\begin{equation}
    e_{\text{block}}(t) = \begin{cases}
        M & z\leq t<T(z)\\
        0 & \text{otherwise}
    \end{cases}, \quad \text{where} \ \int_z^{T(z)} \bar{F}_Y(t) dt = \frac{B\E[Y]}{M\sigma_e}.
\end{equation}
Consistent with Theorem \ref{theorem:front-loading} and \ref{theorem:idling}, the results confirm that concept duration distributions with IMRL benefit from delayed resource allocation up to an optimal delay, whereas for DMRL distributions, introducing delay consistently degrades performance. Overall, these findings demonstrate that the aging properties of training cycles driven by sudden concept drift have a significant impact on the structure of optimal training resource allocation strategies. Optimizing training operations based on the characteristics of concept drift yields substantial performance improvements, particularly under budget-constrained scenarios.
%All executions were executed on a Macbook Air M3 with 16GB of RAM. [MOVED TO THE CAPTION OF FIGURE 2]

\section{Deployment Optimization under Concept Drift}
\label{sec:deployment}

As discussed in Section \ref{sec:system_model}, although improved versions of an ML model may exist on the server side, the performance experienced by clients depends on the version of the model that has been deployed to them. In this section, we study the deployment scheduling problem, aiming to optimize the long-term performance of client-side models subject to a constraint on the deployment rate, which may serve as a proxy for communication costs.

To understand the dynamics governing deployment decisions, we analyze deployment policies in a setting where the server-side model is trained under a fixed allocation of training resources. Following the approach in Section \ref{sec:training}, we restrict our attention to static deployment schedulers, defined as follows:

\begin{definition}[Static Deployment Scheduler]
Let $ D_{i,j}-T_i$
%$\Delta_{i,j} = D_{i,j}-T_i$
denote the deployment time offset relative to concept arrival time $T_i$, where $D_{i,j}$ is the $j$-th deployment time during concept $i$. A deployment scheduler is said to be static -- and hence stationary and causal -- if, for all concepts $i \in \mathbb{N}$,
%$$\Delta_{i,j} = \delta_j, ~\mbox{almost surely}~ \quad \forall i,j \in \mathbb{N},$$
$$  D_{i,j}-T_i = \delta_j, ~\mbox{almost surely}~ \quad \forall i,j \in \mathbb{N},$$
where $\delta_0 = 0 < \delta_1 < \cdots$
%%$\{\delta_{j}:j \in \mathbb{N}\}$ is independent of  both the concept duration $y_i$ and the loss curve %%$g_i(\cdot)$.
so that the set $\{\delta_{j}:j \in \mathbb{N}\}$ defines a static deployment scheduler.
\end{definition}
% Let $N_i=\max\{j:\delta_j<Y_i\}$ be a random variable denoting the number of deployments occurring during concept $i$ under a static deployment scheduler. Assuming a fixed unit training resource allocation for the server-side model, i.e., $e_i(t)= 1, \ \forall t\geq0$, the time-average expected loss as seen by clients under a static deployment scheduler is given by:
% \begin{align}
% \limsup_{t \rightarrow \infty} \frac{1}{t} \int_{\tau=0}^t \hat{\mathcal{L}}(\tau) d\tau = \frac{1}{\E[Y]}\sum_{j=0}^{\infty} \bar{g}(\delta_j)\int_{\delta_j}^{\delta_{j+1}}\bar{F}_Y(t)dt \label{eq:dep_objective}
% \end{align}
% where $\bar{g}(\cdot) =  \mathbb{E}[G_i(\cdot)]$ denotes the expected loss function then  $N_i$ are i.i.d. random variables representing the number of deployments within concept of durations and let  $N \sim N_i$. Similarly, the time-average deployment rate constraint can be expressed as:
% \begin{align}
%     \limsup_{t\rightarrow\infty} \frac{1}{t} \sum_{i=1}^{I(t)}N_i = \frac{\sum_{j=1}^{\infty} \bar{F}_Y(\delta_j)}{\E[Y]}\leq r_D \label{eq:dep_contraint},
% \end{align}
% where $r_D$ denotes the maximum allowable deployment rate. The detailed derivation of Equation \eqref{eq:dep_objective} and \eqref{eq:dep_contraint} is provided in Appendix \ref{supp:time-average}.

Let $N_i=\max\{j:\delta_j<Y_i\}$ denote the random variable representing the number of deployments occurring during concept $i$ under a static deployment scheduler. Assuming a fixed unit training resource allocation for the server-side model, i.e., $e_i(t)= 1, \ \forall t\geq0$, the time-average expected loss experienced by clients under a static deployment scheduler is given by
\begin{align}
    \label{eq:time_average_obj_deployment}
\limsup_{t \rightarrow \infty} \frac{1}{t} \int_{\tau=0}^t \hat{\mathcal{L}}(\tau) d\tau &= \limsup_{t \rightarrow \infty} \frac{1}{t} \int_{\tau=0}^t G_{I(t)}\left(D_{I(t),J(t)}-T_{I(t)}\right) \\
&= \limsup_{t \rightarrow \infty} \frac{1}{t} \left[ \sum_{i=1}^{I(t)-1} \sum_{j=0}^{N_i-1} G_i(\delta_j)(\delta_{j+1}-\delta_j)  +G_i(\delta_{N_i})(Y_i-\delta_{N_i}) \right] \\
&= \frac{\E[\sum_{j=0}^{N_i-1} G_i(\delta_j)(\delta_{j+1}-\delta_j)  +G_i(\delta_{N_i})(Y_i-\delta_{N_i})]}{\E[Y]} \\
&= \frac{\E[\sum_{j=0}^{N-1} \bar{g}(\delta_j)(\delta_{j+1}-\delta_j)  +\bar{g}(\delta_{N})(Y-\delta_{N})]}{\E[Y]} \\
&= \frac{1}{\E[Y]}\sum_{j=0}^{\infty} \bar{g}(\delta_j)\int_{\delta_j}^{\delta_{j+1}}\bar{F}_Y(t)dt, \label{eq:dep_objective-derivation}
\end{align}
where $\bar{g}(\cdot) = \mathbb{E}[G_i(\cdot)]$ denotes the expected loss function, and $N_i$ are i.i.d. random variables representing the number of deployments within each concept durations, with $N \sim N_i$. Similarly, the time-average deployment rate constraint can be expressed as
\begin{align}
    \limsup_{t\rightarrow\infty} \frac{1}{t} \sum_{i=1}^{I(t)}N_i &= \frac{\E [N]}{\E[Y]}\\
  &=\frac{\E[\sum_{j=1}^{\infty} \mathbb{I}\{\delta_j<Y\}]}{\E[Y]} =  \frac{\sum_{j=1}^{\infty} P(\delta_j<Y)}{\E[Y]} \\
  &= \frac{\sum_{j=1}^{\infty} \bar{F}_Y(\delta_j)}{\E[Y]}\leq r_D \label{eq:dep_contraint-derivation},
\end{align}
where $r_D$ denotes the maximum allowable deployment rate. The optimization problem minimizing the time-average expected loss experienced by clients, subject to the time-average deployment rate constraint, is then given by
\begin{align}
\min_{\{\Delta_j\}_{j=1}^{\infty}} \quad & \sum_{j=0}^{\infty} \bar{g}(\delta_j)\int_{\delta_j}^{\delta_{j+1}}\bar{F}_Y(t)dt \label{opt:deployment_problem}\\
\textrm{s.t.} \quad & \Delta_j = \delta_{j}-\delta_{j-1} \geq 0, \ \forall j\in\mathbb{N}^+ \quad \delta_0=0, \\
& \sum_{j=1}^{\infty} \bar{F}_Y(\delta_j) \leq B_2, \quad \text{where } B_2 = {r_D \cdot E[Y]}. \label{eq:deployment_rate_constraint1}
\end{align}
Here ${\{\Delta_j\}_{j=1}^{\infty}}$ denotes the set of inter-deployment durations. Note that the optimization problem \eqref{opt:deployment_problem} is not necessarily convex. However, under suitable conditions, it can be shown to be quasi-convex, as formalized in the following lemma.

\begin{lemma}[Quasi-convexity of Problem \eqref{opt:deployment_problem}]
\label{lemma:quasi-convex_dep}
If $\bar{g}$ is a non-negative, convex, and decreasing function, and $\bar{F}_Y$ is strictly decreasing, then the objective function in \eqref{opt:deployment_problem} is quasi-convex in $\{\Delta_j \geq 0\}_{j \in \mathbb{N}}$. Moreover, if $\bar{F}_Y$ is also convex, then Problem \eqref{opt:deployment_problem} is quasi-convex.
\end{lemma}
A detailed proof of Lemma \ref{lemma:quasi-convex_dep} is provided in Appendix \ref{proof:quasi-convex_dep}. As discussed there, the Karush–Kuhn– Tucker (KKT) conditions are sufficient for global optimality when the feasible set defined by the deployment rate constraint \eqref{eq:deployment_rate_constraint1} is convex. However, even under this condition, a numerical procedure is typically required to obtain the optimal deployment schedule.

Applying the KKT conditions, for a Lagrange multiplier $\nu$ > 0, the optimal deployment scheduler $\{\delta_k^*\}_{k=0}^{\infty}$ satisfies the following necessary condition:
\begin{equation}
\bar{g}'(\delta_k^*)\int_{\delta_k^*}^{\delta_{k+1}^*}\bar{F}_Y(t)dt = \bar{F}_Y(\delta_k^*)[\bar{g}(\delta_k^*) - \bar{g}(\delta_{k-1}^*)] + \nu f_Y(\delta_k^*).
\end{equation}

A numerical approach such as the bisection method can be employed to compute the optimal deployment times, typically by starting from the final deployment and proceeding backward, under an assumed upper bound on the total number of deployments. Motivated by the structure of this solution, we next propose a randomized deployment policy derived from a modified version of \textit{Problem}~\eqref{opt:deployment_problem},
which remains applicable even when the concept duration distribution exhibits a non-convex survival function.

In this modified problem, the deployment rate constraint is replaced by a fixed, deterministic number of deployments, denoted by $N_D$. The resulting optimization problem is given by
\begin{align}
\min_{\{\Delta_j\}_{j=1}^{N_D}} \quad & \sum_{j=0}^{N_D -1} \bar{g} (\delta_j)\int_{\delta_j}^{\delta_{j+1}}\bar{F}_Y(t)dt + \bar{g} (\delta_{N_D})\int_{\delta_{N_D}}^{\infty}\bar{F}_Y(t)dt\label{opt:deployment_problem_relaxed}\\
\textrm{s.t.} \quad & \Delta_j = \delta_{j}-\delta_{j-1} \geq 0, \ \forall j\in[1,N_D] \quad \delta_0=0.
\end{align}
Since the rate constraint has been removed, Lemma~\ref{lemma:quasi-convex_dep} guarantees that Problem \eqref{opt:deployment_problem_relaxed} remains quasi-convex.

\begin{theorem}[Optimal Solution for Problem \eqref{opt:deployment_problem_relaxed}]
\label{theorem:relaxed}
Let $\bar{g}$ be a non-negative, convex, and decreasing function, and let $Y$ be a nonnegative random variable with $\bar{F}_Y(t) > 0$ for all $t \in [0,\infty)$. Then the optimal deployment scheduler $\{\delta_k^*\}_{k=0}^{N_D}$ satisfies
\begin{align}
 \frac{\bar{g}(\delta_{k-1}^*)-\bar{g}(\delta_{k}^*)}{-\bar{g}'(\delta_{k}^*)} = \begin{cases} m_Y(d^*_{N_D}), &k=N_D \\  \frac{\int_{\delta_k^*}^{\delta_{k+1}^*}\bar{F}_Y(t)dt}{\bar{F}_Y(\delta_k^*)}, &k \in [1,N_D-1],
 \end{cases} \label{eq:relaxed_sol}
\end{align}
where $m_Y(\cdot)$ denotes the mean residual life (MRL) function of the concept duration $Y$. Furthermore, the effective deployment rate achieved by the optimal scheduler, defined as $r_e(\{\delta_k^*\}_{k=0}^{N_D})= \sum_{j=1}^{N_D} \bar{F}_Y(\delta_j)$, is monotonically increasing in the number of deployments $N_D$.
% \begin{equation}
%     r_e(\{\delta_k^*\}_{k=0}^{N_D})= \sum_{j=1}^{N_D} \bar{F}_Y(\delta_j),
% \end{equation}
% is monotonically increasing in number of deployments $N_D$.
\end{theorem}
The proof of Theorem \ref{theorem:relaxed} follows directly from the quasi-convexity of Problem \eqref{opt:deployment_problem_relaxed} and the monotonicity of the survival function $\bar{F}_Y(t)$. Full derivations are provided in Appendix~\ref{proof:relaxed}.

In a special case where concept durations follow an exponential distribution and the expected loss decays exponentially, Theorem \ref{theorem:relaxed} admits a closed-form, chain-structured solution that can be determined iteratively.

\begin{corollary}[Exponentially distributed concept durations and Exponential Decaying $\bar{g}$]
\label{corollary:special}
Assume the expected loss function is given by $\bar{g}(t) = \alpha e^{-\beta t}$, with $\alpha, \beta > 0$, and that the concept durations Y are exponentially distributed with rate parameter $\lambda$, i.e., $\bar{F}_Y(t) = e^{-\lambda t}$. Then, the optimal inter-deployment durations $\Delta_k^* = \delta_k^* - \delta_{k-1}^*$ for the problem with a deterministic number of deployment $N_D$ are given by
\begin{equation}
    \Delta_k^* = \begin{cases}
        \frac{1}{\beta} \ln(\frac{\beta}{\lambda}+1) & k=N_D, \\
        \frac{1}{\beta}\ln(\frac{\beta}{\lambda}(1-e^{-\lambda \Delta_{k+1}^*})) &k\in[1,N_{D}-1],
    \end{cases}
    \label{eq:special}
\end{equation}
with $\delta_0=0$ and $k\in[1,N_D]$.
\end{corollary}
The derivation of Corollary \ref{corollary:special} is provided in Appendix \ref{proof:special}. As shown in Equation~\eqref{eq:special}, the final inter-deployment duration is constant, while the preceding durations are determined recursively in a backward fashion, resulting in a chain-structured solution. This structure implies that as the number of deployments $N_D$ increases, the optimal deployment times adjust to accommodate the additional deployments, while previously determined inter-deployment intervals remain unchanged.

% We now introduce a randomized deployment scheduling policy that, at each concept, uses two optimal policies for Problem \eqref{opt:deployment_problem_relaxed} with consecutive number of deployment,  $\{\delta_k^*\}_{k=0}^{N_D}$ and $\{\delta_k^*\}_{k=0}^{N_D+1}$, at random with probability $\gamma$ and $(1-\gamma)$. Such randomized policy achieves an effective deployment rate equals to the convex combinations of $r_e(\{\delta_k^*\}_{k=0}^{N_D})$ and $r_e(\{\delta_k^*\}_{k=0}^{N_D+1})$, which can be matched with the deployment constraint. By leveraging optimal deployment schedulers obtained from Problem \eqref{opt:deployment_problem_relaxed} with different deterministic deployment counts, the randomized deployment scheduler provides a practical solution to the constrained deployment optimization problem \eqref{opt:deployment_problem}. It ensures full utilization of the allowed deployment budget, and does not require the convexity of the survival function. Our simulation results further demonstrate that the performance of the randomized scheduler closely approximates that of the optimal (non-randomized) policy, making it an effective choice for deployment under concept drift. The details of the randomized scheduler is discussed in Appendix \ref{supp:randomized}.

We now introduce a randomized deployment scheduling policy to address the deployment optimization problem under the rate constraint \eqref{opt:deployment_problem}. The proposed approach leverages the optimal deterministic schedulers obtained from Problem \eqref{opt:deployment_problem_relaxed} for different deployment counts. The randomized policy operates over the convex hull of these optimal schedulers with consecutive numbers of deployments, assigning probabilities such that the resulting expected effective deployment rate exactly matches the allotted deployment rate limit.

\begin{theorem}[Randomized Deployment Scheduler]
\label{theorem:randomized}
Let $\{\delta_k^*\}_{k=0}^{\infty}$ denote the optimal deployment scheduler for Problem \eqref{opt:deployment_problem}, and let $\{\delta_k^*\}_{k=0}^{N_D}$ denote the optimal deployment scheduler for Problem \eqref{opt:deployment_problem_relaxed} with a deterministic number of deployments $N_D$. Assume that the expected loss function $\bar{g}(\cdot)$ is non-negative, convex, and decreasing, and that the concept duration $Y$ has survival function $\bar{F}_Y(t)> 0$ for all $t\in[0,\infty)$. Then, for any optimal scheduler $\{\delta_k^*\}_{k=0}^{\infty}$, there exists a deployment count $N_D^*\in\mathbb{N}^+$ and a scalar parameter $\gamma \in [0,1]$ such that a convex combination of the effective deployment rates corresponding to the optimal schedulers with $N_D^*$ and $N_D^* + 1$ deployments exactly matches the allotted deployment rate limit, i.e.,
\begin{equation}
    \gamma \cdot r_e( \{\delta_k^*\}_{k=0}^{N_D^*}) + (1-\gamma) \cdot r_e( \{\delta_k^*\}_{k=0}^{N_D^*+1}) = r_e(\{\delta_k^*\}_{k=0}^{\infty}) = r_D \E[Y],
\end{equation}
where $r_D$ denotes the allotted deployment rate limit.

A randomized deployment policy that, at the beginning of each concept, uses the scheduler $\{\delta_k^*\}_{k=0}^{N_D^*} $ with probability $\gamma$ and $\{\delta_k^*\}_{k=0}^{N_D^*+1}$ with probability $1 - \gamma$ achieves an expected effective deployment rate exactly equal to the deployment rate constraint.
\end{theorem}
A detailed proof of Theorem \ref{theorem:randomized} is presented in Appendix \ref{proof:randomized}.

The randomized deployment scheduler provides an approximate solution to the constrained deployment optimization problem \eqref{opt:deployment_problem}, which is inherently non-convex. It guarantees full utilization of the allotted deployment budget and does not rely on the convexity of the survival function of the concept duration distribution. Although the method requires knowledge of the expected loss curve and the probability distribution of concept duration, it offers a computationally efficient alternative to high-dimensional grid search or unstable non-convex optimization techniques. While the randomized policy assumes that both the expected loss curve and concept duration distribution can be estimated -- an assumption that is feasible in many practical scenarios -- an alternative approach is to use a parametric scheduler that emulates the key structural property of the optimal solution, namely, decreasing inter-deployment durations. In this context, the randomized policy serves as a computationally efficient baseline for testing and tuning such parametric algorithms.

Our simulation results further demonstrate that the performance of the randomized scheduler closely approximates that of the optimal (non-randomized) policy,
making it an effective and computationally efficient choice for deployment under concept drift.
\begin{figure}[h]
    \centering
    \begin{subfigure}{0.32\linewidth}        \includegraphics[width=\linewidth]{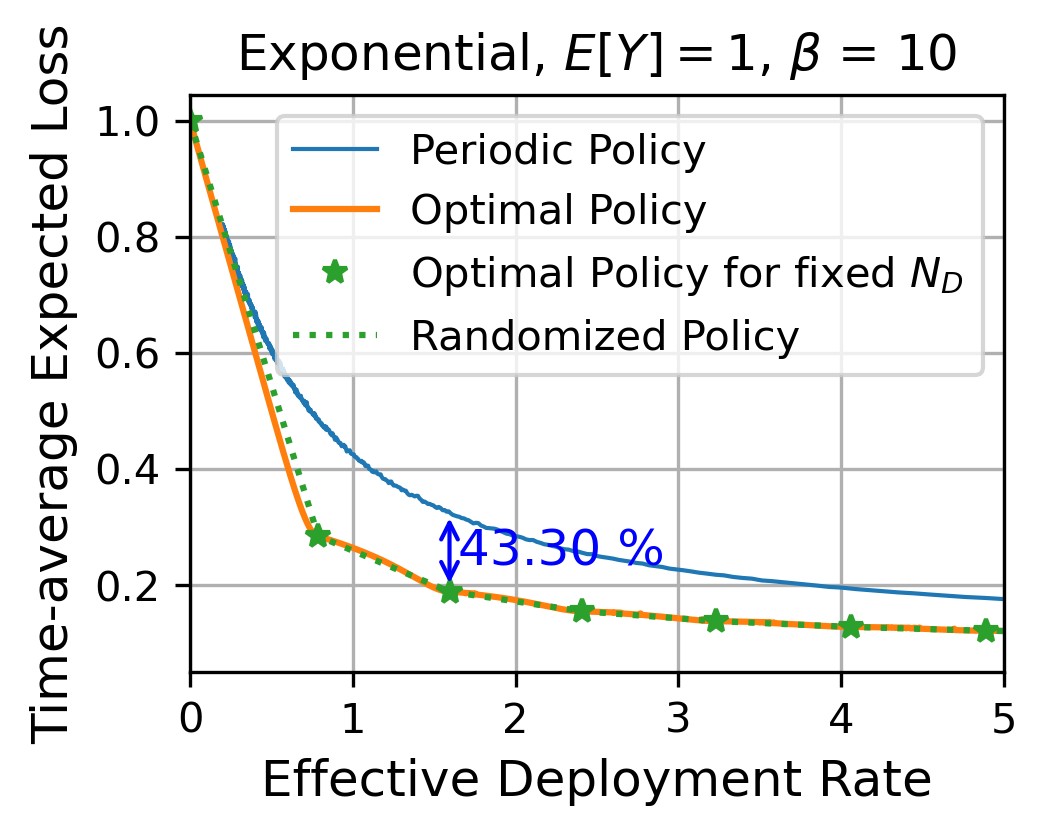}
    \end{subfigure}
    \hfill
    \begin{subfigure}{0.32\linewidth}
    \includegraphics[width=\linewidth]{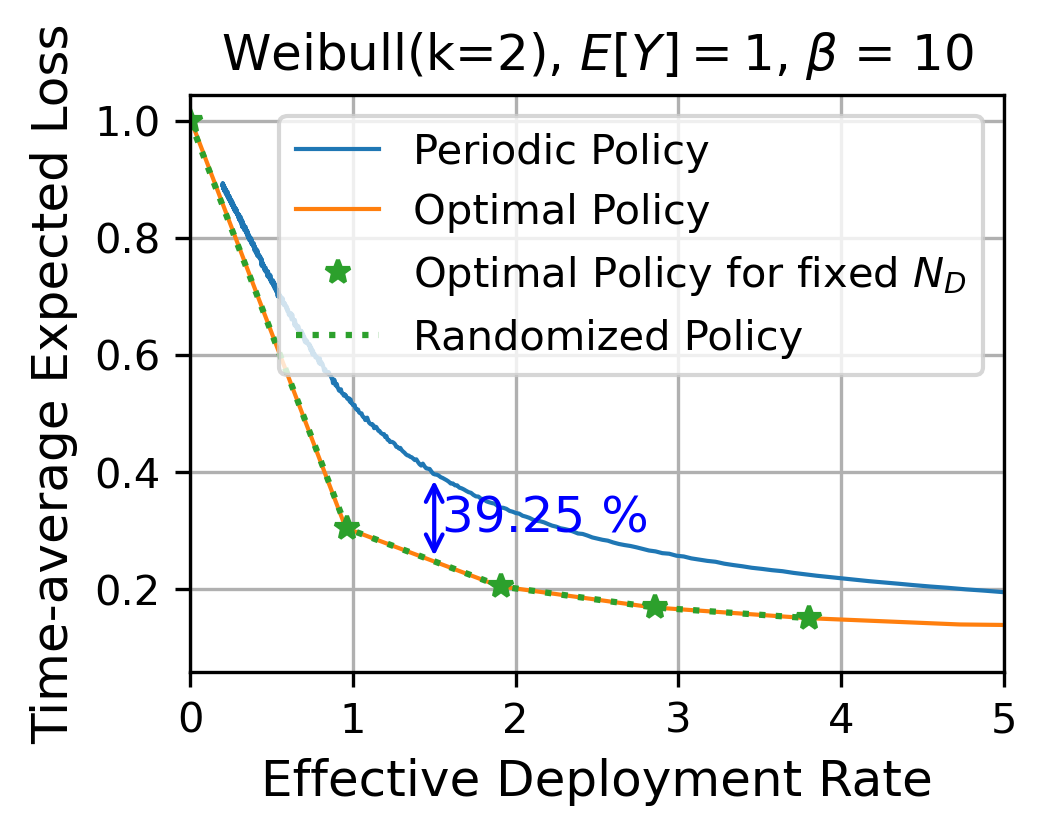}
    \end{subfigure}
    \hfill
    \begin{subfigure}{0.32\linewidth}
        \includegraphics[width=\linewidth]{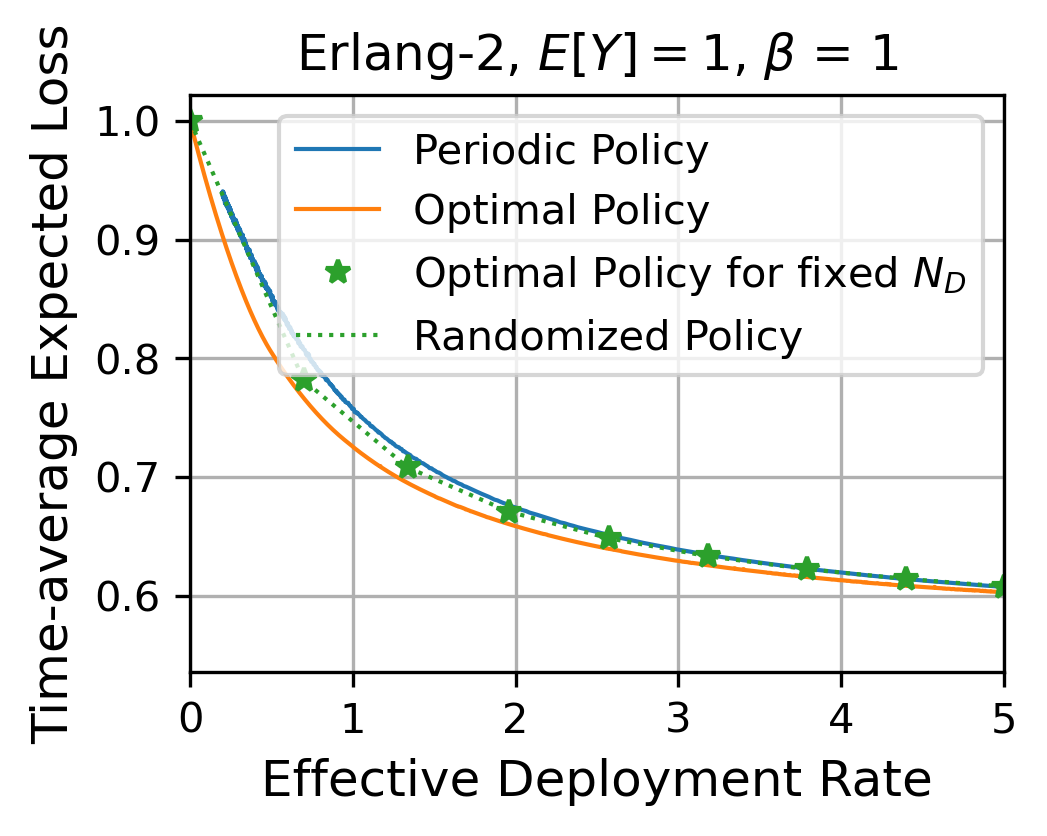}
    \end{subfigure}
    \caption{Comparison of periodic, optimal, and randomized deployment policies under different concept duration distributions
    (Exponential, Weibull($k=2$), and Erlang-2), with $\mathbb{E}[Y]=1$ and exponentially decaying expected loss $\bar{g}(t)=e^{-\beta t}$.
    Simulations were run on a MacBook Air M3 (16 GB RAM).}
    \label{fig:deployment}
\end{figure}
Figure \ref{fig:deployment} presents simulation results highlighting the performance gains achieved by optimal deployment policies relative to a periodic deployment baseline and demonstrates the near-optimality of the randomized deployment policy across different concept duration distributions. In all experiments, the expected loss follows an exponential decay, and distribution parameters are chosen such that $\mathbb{E}[Y]=1$. Except for the Exponential case, where a closed-form solution is available, the optimal deployment policies are computed numerically. For Exponential and Weibull($k=2$) distributions, the optimal policy reduces the client-side time-average expected loss by up to 43.30\% and 39.25\%, respectively, compared to the periodic policy under the same effective deployment rate. In these settings, the randomized policy, constructed as a convex combination of optimal solutions for adjacent deployment counts $N_D$, achieves performance nearly indistinguishable from the optimal policy. In contrast, for the Erlang-2 distribution, the performance gap between optimal and periodic policies narrows, owing both to the distribution's reduced variability (which limits the policy's leverage) and to the smaller decay rate $\beta$ of the loss function, which slows the decline of expected penalties. Under these conditions, the randomized policy no longer matches the optimal performance, providing a concrete example where the optimal policy strictly outperforms its randomized counterpart.

\section{Conclusion}

We have presented a formal framework for studying training resource allocation and model deployment in distributed machine learning systems operating under persistent, discrete concept drift. In scenarios where training resources are elastic, i.e., available on demand but incurring computational and communication costs, a provider must optimize when to allocate resources and when to deploy updated models. Our analysis shows that the structure of optimal policies depends fundamentally on the statistical properties of concept durations and loss functions. As machine learning systems become increasingly ubiquitous, the monitoring-retraining-redeployment cycle will represent a growing component of their operational costs. The results of this work lay the foundation for principled, cost-efficient training and deployment strategies that adapt to non-stationary environments.

\newpage
\bibliographystyle{ACM-Reference-Format}
\bibliography{references}
\newpage
\appendix

\section{Technical Appendices}

\subsection{Proof of Lemma \ref{lem:convexity}}
\label{proof:lem:convexity}
\begin{proof}
Consider any $e_1,e_2 \in \mathcal{E}$ and any $\lambda \in [0,1]$. It holds that
\begin{align}
    J(\lambda e_1 + (1-\lambda)e_2) =& \int_0^{\infty} \bar{g}\left(\int_0^{t}(\lambda e_1(\tau) + (1-\lambda)e_2(\tau))d\tau\right) \bar{F}_Y(t)dt\\
    \leq& \int_0^{\infty} \left(\lambda \bar{g}\left(\int_0^{t}e_1(\tau) d\tau\right) + (1-\lambda) \bar{g}\left(\int_0^{t} e_2(\tau) d\tau\right) \right)\bar{F}_Y(t)dt\label{eq:convexity_h1}\\
   =& \lambda J(e_1) + (1-\lambda)J(e_2),
\end{align}
where the inequality \eqref{eq:convexity_h1} follows from the fact that $\bar{g}(\cdot)$ is a convex function. Therefore, $J(e)$ is a convex functional on the domain $\mathcal{E}$.
Similarly, the functional $C(e)$ is affine since
\begin{align}
    C(\lambda e_1 + (1-\lambda) e_2) &= \int_0^{\infty} \left(\lambda e_1(t) + (1-\lambda) e_2(t)\right) \bar{F}_Y(t) dt \\
    &= \lambda C(e_1) + (1-\lambda) C(e_2),
\end{align}
which completes the proof.
\end{proof}

\subsection{Proof of Theorem \ref{theorem:front-loading}}
\label{proof:front-loading}
\begin{proof}
    For the front-loading policy to be optimal, we need to show that the switching function crosses zero at most once from negative to positive.
    Suppose $0 < B < M\sigma_e$. Here, $t^*$ is uniquely determined by the budget constraint \eqref{eq:reformulated_constraint1}, with $0 < t^*<\infty$. Since the budget constraint is binding, by the complementary slackness \eqref{eq:complementary_slackness_main}, we expect $\nu>0$. Given the policy \eqref{eq:opt_res_alloc}, the state and costate trajectories are
    \begin{align}
    x^*(t) &= \begin{cases} Mt & t<t^* \\
Mt^* & t \geq t^*
\end{cases} \\
p^*(t) &= \begin{cases} \int_t^{t^*}\bar{g}'(Ms)\bar{F}_Y(s)ds + \bar{g}'(Mt^*)m_Y(t^*)\bar{F}_Y(t^*) & t<t^* \\
\bar{g}'(Mt^*)m_Y(t)\bar{F}_Y(t) & t \geq t^*
\end{cases}
\end{align}
At the switch point $t^*$, we require $\phi(t^*) = 0$, i.e.,
\begin{equation}
    \phi(t^*) = p^*(t^*) + \nu \bar{F}_Y(t^*) = \bar{F}_Y(t^*)(\bar{g}'(Mt^*)m_Y(t^*) + \nu) = 0.
\end{equation}
Since $\bar{F}_Y(t) >0$ and $\bar{g}'(t) < 0$, for $0<t<\infty$, we have $\nu > 0$ and
\begin{equation}
     \nu = -\bar{g}'(Mt^*)m_Y(t^*),
\end{equation}
which is consistent with \eqref{eq:complementary_slackness_main}.

Now, we verify \eqref{eq:control_opt_main} by showing $\phi(t) > 0$ for $t \in (t^*,\infty)$, and $\phi(t) < 0$ for $t \in [0,t^*)$.
For $t \in (t^*,\infty)$, where $e^*(t) = 0$, the switching function becomes
\begin{align}
    \phi(t) &= p^*(t) + \nu \bar{F}_Y(t) = \bar{g}'(Mt^*)m_Y(t)\bar{F}_Y(t) + (-\bar{g}'(Mt^*)m_Y(t^*))\bar{F}_Y(t) \\
    &=\bar{F}_Y(t) \bar{g}'(Mt^*) [m_Y(t) - m_Y(t^*)].
\end{align}
Since $\bar{F}_Y(t) \ge 0$ and $\bar{g}'(Mt^*) < 0$, for $\phi(t) > 0$ we need $m_Y(t) < m_Y(t^*)$. Since $t > t^*$, the condition $m_Y(t) < m_Y(t^*)$ holds if $m_Y(t)$ is decreasing for $t > t^*$.

For $t \in [0,t^*)$, where $e^*(t) = M$, the first derivative of the switching function is
\begin{equation}
    \phi'(t) = \bar{F}_Y(t)[-\bar{g}'(x^*(t)) - \nu h_Y(t)] = \bar{F}_Y(t)[-\bar{g}'(Mt) + \bar{g}'(Mt^*)m_Y(t^*)h_Y(t)].
\end{equation}
Since $\bar{g}'(Mt)<\bar{g}'(Mt^*)$ for $t<t^*$,
\begin{equation}
    -\bar{g}'(Mt) + \bar{g}'(Mt^*)m_Y(t^*)h_Y(t) > -\bar{g}'(Mt^*)(1-m_Y(t^*)h_Y(t)).
\end{equation}
If $m_Y(t) > m_Y(t^*)$ for $t < t^*$, then
$$1-m_Y(t^*)h_Y(t)>1-m_Y(t)h_Y(t)>0,$$
which since $\bar{F}_Y(t)>0$ means $\phi'(t)>0$ for $t \in [0,t^*)$. Given that $\phi(t^*) = 0$, this implies $\phi(t)<0$ for $t \in [0,t^*)$.

For the case  $B \geq M \sigma_e $, since the budget constraint is not binding, by \eqref{eq:complementary_slackness_main}, $\nu =0$. Since $e^*(t) = M$ for $t \in [0,\infty)$, the state function satisfies $x^*(t) = Mt$, the switching function $\phi(t) = p(t)$, and the derivative of the costate $p'(t) = -\bar{g}'(Mt)\bar{F}_Y(t)$. Since $\bar{g}'(t) < 0$ and $\bar{F}_Y(t)>0$, $p'(t) > 0$ for $t \in [0,\infty)$. Since by the transversality condition \eqref{eq:transversality_main} $\lim_{t\rightarrow\infty}p(t) = 0$ and $p(t)$ is increasing, we conclude $\phi(t) = p(t) < 0$.
\end{proof}

\subsection{Proof of Theorem \ref{theorem:idling}}
\label{proof:idling}
\begin{proof}
Suppose there is an optimal resource allocation policy that idles until time $t^*$, i.e., $e^*(t) = 0, \ \forall t\in [0,t^*]$, and thus $x(t) = 0, \ \forall t \in [0,t^*]$.

At the switch point $t^*$ we require $\phi(t^*) = 0$,
\begin{equation}
    \phi(t^*) = p^*(t^*) + \nu \bar{F}_Y(t^*) = 0 . \label{eq:nu_backloading_0}
\end{equation}
Since the resource budget is binding, i.e., $B<M\sigma_e$, by \eqref{eq:nu_backloading_0} and Mean Value Theorem there exists $x_0$ such that $0\leq x_0<\infty$, which yields
\begin{align}
    \nu &= -\frac{p^*(t^*)}{\bar{F}_Y(t^*)} = -\frac{\int_{t^*}^{\infty}\bar{g}'(x^*(s))\bar{F}_Y(s)ds}{\bar{F}_Y(t^*)} = -\frac{\bar{g}'(x_0)\int_{t^*}^{\infty}\bar{F}_Y(s)ds}{\bar{F}_Y(t^*)} \\
    &= -\bar{g}'(x_0)m_Y(t^*) > 0,
\end{align}
where \eqref{eq:complementary_slackness_main} is satisfied since $\bar{g}'(t)<0$.

By \eqref{eq:control_opt_main} and \eqref{eq:control_law_main_final}, for $t \in [0,t^*)$ we have $\phi(t) > 0$. Since $\phi(t^*) = 0$, the first derivative of the switching function for $t \in [0,t^*)$ should be positive, i.e.,
\begin{equation}
    \phi'(t) = \bar{F}_Y(t)[-\bar{g}'(x^*(t)) - \nu h_Y(t)] = \bar{F}_Y(t)[-\bar{g}'(0) + \bar{g}'(x_0)m_Y(t^*)h_Y(t)] > 0.
\end{equation}
Since $\bar{g}'(0)\leq\bar{g}'(x_0)$ for $0\leq x_1 < \infty$,
\begin{equation}
    -\bar{g}'(0) + \bar{g}'(x_0)m_Y(t^*)h_Y(t) \geq -\bar{g}'(x_0)(1-m_Y(t^*)h_Y(t)).
\end{equation}
If $m_Y(t) > m_Y(t^*)$ for $t < t^*$, then
$$1-m_Y(t^*)h_Y(t)>1-m_Y(t)h_Y(t)>0,$$
where the last inequality follows from the IMRL property. Thus, since $\bar{F}_Y(t)>0$, if $Y$ has IMRL, then $\phi'(t)>0$ for $t \in [0,t^*)$.
\end{proof}

\subsection{Proof of Corollary \ref{corollary:back-loading}}
\label{proof:back-loading}
\begin{proof}
    For the back-loading policy to be optimal, we need to show that the switching function crosses zero at most once from positive to negative.
    Suppose $0 < B < M\sigma_e$. Here, $t^*$ is uniquely determined by the budget constraint \eqref{eq:reformulated_constraint1}, with $0 < t^*<\infty$. Since the budget constraint is binding, by the complementary slackness \eqref{eq:complementary_slackness_main}, we expect $\nu>0$. Given the policy \eqref{eq:opt_res_alloc}, the state and costate trajectories are
    \begin{align}
    x^*(t) &= \begin{cases} 0 & t<t^* \\
M(t-t^*) & t \geq t^*
\end{cases} \\
p^*(t) &= \begin{cases} \int_{t^*}^{\infty}\bar{g}'(M(s-t^*))\bar{F}_Y(s)ds  & t<t^* \\
\int_{t}^{\infty}\bar{g}'(M(s-t^*))\bar{F}_Y(s)ds & t \geq t^*.
\end{cases}
\end{align}
At the switch point $t^*$ we require $\phi(t^*) = 0$,
\begin{equation}
    \phi(t^*) = p^*(t^*) + \nu \bar{F}_Y(t^*) = 0 . \label{eq:nu_backloading}
\end{equation}
By \eqref{eq:nu_backloading} and Mean Value Theorem, there exists $x_1$ such that $0\leq x_1<\infty$ which yields
\begin{align}
    \nu &= -\frac{p^*(t^*)}{\bar{F}_Y(t^*)} = -\frac{\int_{t^*}^{\infty}\bar{g}'(M(s-t^*))\bar{F}_Y(s)ds}{\bar{F}_Y(t^*)} = -\frac{\bar{g}'(x_1)\int_{t^*}^{\infty}\bar{F}_Y(s)ds}{\bar{F}_Y(t^*)} \\
    &= -\bar{g}'(x_1)m_Y(t^*) > 0,
\end{align}
where \eqref{eq:complementary_slackness_main} is satisfied since $\bar{g}'(t)<0$.

Now, we verify \eqref{eq:control_opt_main} by showing $\phi(t) > 0$ for $t \in [0,t^*)$, and $\phi(t) < 0$ for $t \in (t^*,\infty)$.
For $t \in [0,t^*)$, where $e^*(t) = 0$, the first derivative of the switching function is
\begin{equation}
    \phi'(t) = \bar{F}_Y(t)[-\bar{g}'(x^*(t)) - \nu h_Y(t)] = \bar{F}_Y(t)[-\bar{g}'(0) + \bar{g}'(x_1)m_Y(t^*)h_Y(t)].
\end{equation}
Since $\bar{g}'(0)\leq\bar{g}'(x_1)$ for $0\leq x_1 < \infty$,
\begin{equation}
    -\bar{g}'(0) + \bar{g}'(x_1)m_Y(t^*)h_Y(t) \geq -\bar{g}'(x_1)(1-m_Y(t^*)h_Y(t)).
\end{equation}
If $m_Y(t) > m_Y(t^*)$ for $t < t^*$, then
$$1-m_Y(t^*)h_Y(t)>1-m_Y(t)h_Y(t)>0,$$
which since $\bar{F}_Y(t)>0$ means $\phi'(t)>0$ for $t \in [0,t^*)$, and the last inequality follows from the IMRL property. Given that $\phi(t^*) = 0$, this implies $\phi(t)<0$ for $t \in [0,t^*)$.

For $t \in (t^*,\infty)$, where $e^*(t) = M$, the switching function becomes
\begin{align}
    \phi(t) &= \int_{t}^{\infty}\bar{g}'(M(s-t^*))\bar{F}_Y(s)ds - \frac{\int_{t^*}^{\infty}\bar{g}'(M(s-t^*))\bar{F}_Y(s)ds}{\bar{F}_Y(t^*)}\bar{F}_Y(t).
    \label{eq_phi_back}
\end{align}
The condition $\phi(t) < 0$ for $t \in (t^*,\infty)$ is satisfied if the following inequality holds for $t\in (t^*,\infty)$:
\begin{equation}
    \frac{\int_{t}^{\infty}\bar{g}'(M(s-t^*))\bar{F}_Y(s)ds}{\bar{F}_Y(t) } < \frac{\int_{t^*}^{\infty}\bar{g}'(M(s-t^*))\bar{F}_Y(s)ds}{\bar{F}_Y(t^*)}.
    \label{eq:cond_back}
\end{equation}
For the expected loss function with constant negative first derivative, i.e., $\bar{g}'(t) = -\beta, \ \beta>0$, Equation \eqref{eq:cond_back} becomes
\begin{equation}
    m_Y(t) > m_Y(t^*), \quad \forall t \in (t^*,\infty).
\end{equation}

For the case $B \geq M \sigma_e $, since the budget constraint is not binding, by \eqref{eq:complementary_slackness_main}, $\nu =0$. Since $e^*(t) = M$ for $t \in [0,\infty)$, the state function is $x^*(t) = Mt$, the switching function is $\phi(t) = p(t)$, and the derivative of the costate is $p'(t) = -\bar{g}'(Mt)\bar{F}_Y(t)$. Since $\bar{g}'(t) < 0$ and $\bar{F}_Y(t)>0$, $p'(t) > 0$ for $t \in [0,\infty)$. Since due to the transversality condition \eqref{eq:transversality_main} $\lim_{t\rightarrow\infty}p(t) = 0$ and $p(t)$ is increasing, $\phi(t) = p(t) < 0$.
\end{proof}

\subsection{Proof of Lemma \ref{lemma:quasi-convex_dep}}
\label{proof:quasi-convex_dep}
\begin{proof}
Let us define a step function $s: [0,\infty) \rightarrow \mathbb{R}$ associated with the deployment offsets $\{\delta_j\}_{j=0}^{K}$ as
\begin{equation}
    s(t) = \begin{cases}
        \delta_j & \text{if }t \in [\delta_j, \delta_{j+1}) \text{ for } j=0, \dots, K-1, \\
        \delta_K & \text{if }t \in [\delta_K, \infty),
    \end{cases} \label{eq:step}
\end{equation}
where $K = \sup\{k:k\in \mathbb{N}\}$, $\delta_{K+1} = \infty$ and $\delta_0=0$. Due to \eqref{eq:step}, the objective function \eqref{opt:deployment_problem} can be written as a single integral,
\begin{equation}
    h(\{\delta_j\}_{j=1}^{K}) = \int_0^{\infty} \bar{g}(s(t))\bar{F}_Y(t)dt.
\end{equation}
Let $\{\delta_j^{(1)}\}_{j=1}^{K}$ and $\{\delta_j^{(2)}\}_{j=1}^{K}$ be two different set of offsets, and $z_j = \lambda \delta_j^{(1)} + (1-\lambda) \delta_j^{(2)}, \ \lambda \in [0,1] $ be the component-wise convex combination. Then $\{z_j\}_{j=1}^{K}$ also satisfies the ordering constraint and is feasible. Let $s^{(1)}(t),\ s^{(2)}(t),\ s^{(z)}(t)$ be the corresponding step functions.

Let $i^{(1)}$, $i^{(2)}$ be such that $t \in [\delta_{i^{(1)}}^{(1)},\delta_{i^{(1)}+1}^{(1)})$ and $t \in [\delta_{i^{(2)}}^{(2)},\delta_{i^{(2)}+1}^{(2)})$, respectively. Let $m = \min\{i^{(1)},i^{(2)}\}$. Since $\delta_m^{(1)},\delta_m^{(2)} \leq t$, their convex combination $z_m \leq t$. Hence, the left endpoint of the interval in which $t$ falls under $\{z_j\}_{j=1}^{K}$ is at most $t$, implying that $s^{(z)}(t)\geq z_m \geq \min\{\delta^{(1)}_m,\delta^{(2)}_m\}$. Thus,
\begin{equation}
    s^{(z)}(t) \geq \min\{s^{(1)}(t),s^{(2)}(t)\} \label{eq:step_quasi}.
\end{equation}
Since $\bar{g}$ is non-increasing, $\bar{F}_Y(t)\geq 0 , \ \forall t $, and by \eqref{eq:step_quasi} it follows that for any fixed $t$
\begin{align}
    \bar{g}(s^{(z)}(t))F_Y(t)&\leq \bar{g}(\min\{s^{(1)}(t),s^{(2)}(t)\})F_Y(t) \\
    &\leq \max\{ \bar{g}(s^{(1)}(t))F_Y(t), \bar{g}(s^{(2)}(t))F_Y(t)\}.
\end{align}
Thus, the function $\{\delta_j\}_{j=0}^{K} \mapsto \bar{g}(s(t))F_Y(t)$ is quasi-convex in $\{\delta_j\}_{j=0}^{K}$ for any fixed $t\in[0,\infty)$.

Under the assumption that the offset sets $\{\delta_j^{(1)}\}_{j=1}^{K}$ and $\{\delta_j^{(2)}\}_{j=1}^{K}$ are component-wise greater or equal, i.e., $\delta_j^{(1)} \geq \delta_j^{(2)}, \ \forall j$, since the pointwise dominance of either of the offset preserved for all $t$, it follows that
\begin{equation}
    \int_0^{\infty} \bar{g}(s^{(z)}(t))F_Y(t) dt \leq \max\{ \int_0^{\infty} \bar{g}(s^{(1)}(t))F_Y(t) dt,  \int_0^{\infty} \bar{g}(s^{(2)}(t))F_Y(t) dt\}.
\end{equation}
\end{proof}

\subsection{Proof of Theorem \ref{theorem:relaxed}}
\label{proof:relaxed}
\begin{proof}
Given the objective function \eqref{opt:deployment_problem_relaxed},
\begin{equation}
    h(\{\delta_j\}_{j=1}^{N_D}) := \sum_{j=0}^{N_D-1} \bar{g} (\delta_j)\int_{\delta_j}^{\delta_{j+1}}\bar{F}_Y(t)dt + \bar{g} (\delta_{N_D})\int_{\delta_{N_D}}^{\infty}\bar{F}_Y(t)dt.
\end{equation}
By the first order optimality conditions, we have
\begin{align}
    \left. \frac{\partial h}{\partial \delta_{N_D}}\right |_{\{\delta_j^*\}_{j=1}^{N_D}} &= \bar{g} (\delta^*_{N_D-1}) \bar{F}_Y(\delta^*_{N_D}) + \bar{g}' (\delta^*_{N_D}) \int_{\delta^*_{N_D}}^{\infty}\bar{F}_Y(t)dt  -   \bar{g} (\delta^*_{N_D}) \bar{F}_Y(\delta^*_{N_D}) = 0, \\
\end{align}
and for $j=1,\ldots N_D-1$ it holds that
    \begin{align}
    \left. \frac{\partial h}{\partial \delta_{j}}\right |_{\{\delta_j^*\}_{j=1}^{N_D}} &= \bar{g} (\delta^*_{j-1}) \bar{F}_Y(\delta^*_{j}) +\bar{g}' (\delta^*_{j}) \int^{\delta^*_{j}}_{\delta^*_{j-1}} \bar{F}_Y(t)dt   -   \bar{g} (\delta^*_{j}) \bar{F}_Y(\delta^*_{j}) = 0 .\\
\end{align}
Rearranging these terms yields \eqref{eq:relaxed_sol}.

To prove that the effective rate is increasing in $N_D$, let $\delta_j^*(N_D)$ denote the $j$-th optimal deployment offset where $N_D$ is the total number of deployments. By the convexity of $\bar{g}$, the monotonicity of $\bar{F}_Y$, and the KKT conditions, one can show that
\begin{equation}
    \delta_j^*(N_D+1) < \delta_j^*(N_D), \quad \text{for} \ j=1,\cdots, N_D. \label{eq:key_component}
\end{equation}
Since $\bar{F}_Y$ is decreasing, \eqref{eq:key_component} implies that
\begin{equation}
   \bar{F}_Y (\delta_j^*(N_D+1)) > \bar{F}_Y (\delta_j^*(N_D)) , \quad \text{for} \ j=1,\cdots, N_D.
\end{equation}
Therefore, the difference between $r_e(\{\delta_j^*\}_{j=0}^{N_D+1}) $ and $r_e(\{\delta_j^*\}_{j=0}^{N_D}) $ is positive, i.e.,
\begin{align}
    r_e(\{\delta_j^*\}_{j=0}^{N_D+1}) - r_e(\{\delta_j^*\}_{j=0}^{N_D}) =& \sum_{j=1}^{N_D+1}  \bar{F}_Y (\delta_j^*(N_D+1)) - \sum_{j=1}^{N_D}  \bar{F}_Y (\delta_j^*(N_D)) \\
    =& \bar{F}_Y (\delta_{N_D+1}^*(N_D+1)) + \sum_{j=1}^{N_D}  \bar{F}_Y (\delta_j^*(N_D+1)) \\
    &- \bar{F}_Y (\delta_j^*(N_D)) > 0.
\end{align}

\end{proof}

\subsection{Proof of Corollary \ref{corollary:special}}
\label{proof:special}
\begin{proof}
For $\bar{g}(t) = \alpha e^{-\beta t}$ and $\bar{F}_Y(t) = e^{-\lambda t}$, by Theorem \ref{theorem:relaxed} it holds that
\begin{align}
   \frac{1}{\beta} e^{\beta \Delta_{k}^*}(1-e^{-\beta \Delta_k^*}) =
    \begin{cases}
        1 & k=N_D\\
        \frac{1}{\lambda} (1-e^{-\lambda\Delta_{k+1}^*}) &k = 1,\cdots,N_D-1.
    \end{cases}
\end{align}
Rearranging the terms yields \eqref{eq:special}.
\end{proof}

\subsection{Proof of Theorem \ref{theorem:randomized}}
\label{proof:randomized}
\begin{proof}
Under the randomized policy described in Theorem \ref{theorem:randomized}, the number of deployments within concept durations depends on whether the deployment schedule in concept $i$ is $\pi_i =  \{\delta_k^*\}_{k=0}^{N_D^*}$ or $\pi_i =  \{\delta_k^*\}_{k=0}^{{N_D+1}^*}$. Therefore, the time-average deployment rate under randomized deployment policy can be written as
\begin{align}
    \limsup_{t\rightarrow\infty} \frac{1}{t} \sum_{i=1}^{I(t)}N_i &= \frac{\E [N \cdot \mathbb{I}\{\pi =  \{\delta_k^*\}_{k=0}^{N_D^*}\} ] + \E [N \cdot \mathbb{I}\{\pi =  \{\delta_k^*\}_{k=0}^{N_D+1^*}\} ]}{\E[Y]}\\
    &= \frac{\text{P}(\pi =  \{\delta_k^*\}_{k=0}^{N_D^*}) \sum_{j=1}^{N_D^*} \bar{F}_Y(\delta_j)  + \text{P}(\pi =  \{\delta_k^*\}_{k=0}^{N_D+1^*}) \sum_{j=1}^{N_D+1^*} \bar{F}_Y(\delta_j)}{\E[Y]} \label{eq:2}\\
    &= \frac{\gamma r_e( \{\delta_k^*\}_{k=0}^{N_D^*})  + (1-\gamma) r_e( \{\delta_k^*\}_{k=0}^{N_D+1^*})}{\E[Y]} \label{eq:eff_time-ave},
\end{align}
where \eqref{eq:2} follows from the independence of the number of deployments and the fixed deployment policy selections in each concept.

As shown in Theorem \ref{theorem:relaxed}, since the effective deployment rate achieved by the optimal scheduler $r_e(\{\delta_k^*\}_{k=0}^{N_D})= \sum_{j=1}^{N_D} \bar{F}_Y(\delta_j)$ is monotonically increasing in the number of deployments $N_D$ within the range $[0,\infty)$, there exists $N_D^* \in \mathbb{N}$ and $\gamma \in [0,1]$ which ensure that the time-average deployment rate in \eqref{eq:eff_time-ave} matches the deployment rate limit $r_D$.
\end{proof}

\subsection{Additional Simulation Results with Alternative Expected Loss Curves}
\label{app:additional_simulations}
Note that our analysis of the optimal policies applies to any convex decreasing expected loss curve. However, the improvements over intuitive deployment heuristics depend on the probability distribution of concept duration, functional form of the expected loss function, and additional system parameters. In the figures below, we provide additional simulation results for the training resource allocation problem with different expected loss curves.
\begin{figure}[h]
    \centering
    \begin{subfigure}{0.48\linewidth}
    \centering
        \includegraphics[height=4.5cm]{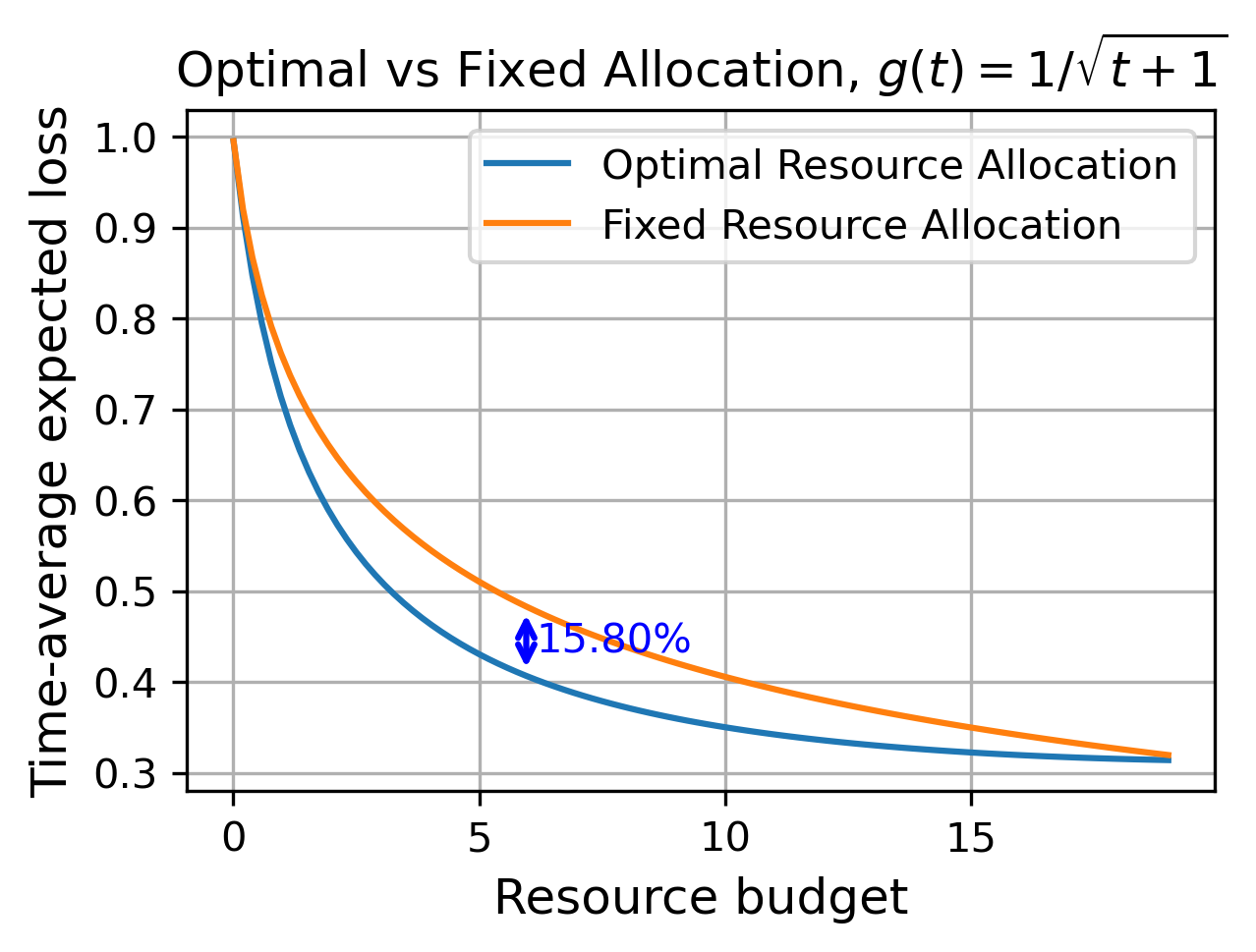}
        \caption{Comparison between constraint-binding optimal and fixed resource allocation policies. $Y\sim \text{Exp}(1)$, $\bar{g}(t)=1/\sqrt{1+t}$, $ \sigma_e=1,M=20$.}
        \label{fig:add_sim_sqrt}
    \end{subfigure}
    \hfill
    \begin{subfigure}{0.48\linewidth}
    \centering
        \includegraphics[height=4.5cm]{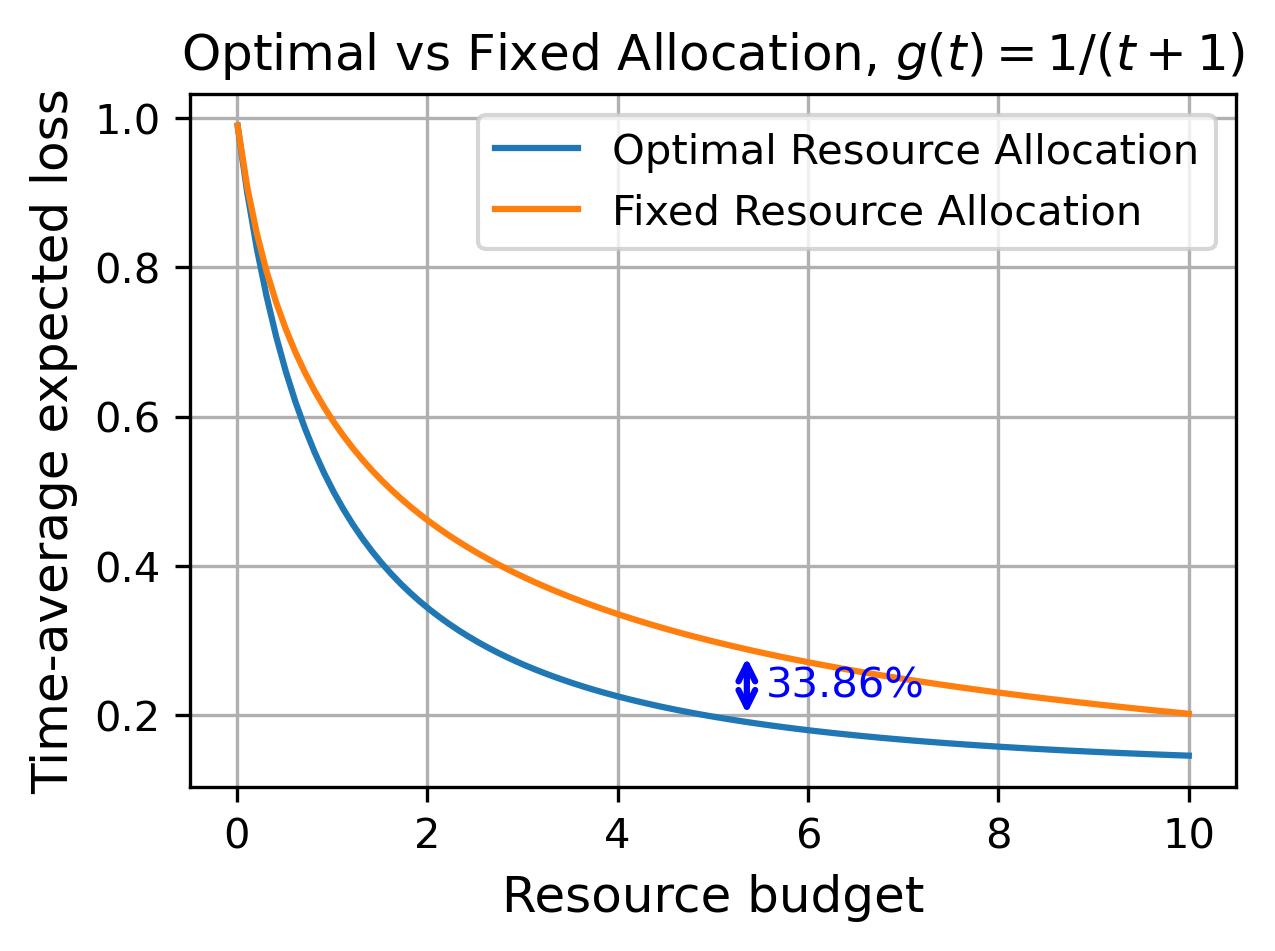}
        \caption{Comparison between constraint-binding optimal and fixed resource allocation policies. $Y\sim \text{Exp}(1)$, $\bar{g}(t)=1/(1+t)$, $ \sigma_e=1,M=20$.}
        \label{fig:add_sim_sqrt}
    \end{subfigure}
    \vskip\baselineskip
    \begin{subfigure}{0.48\linewidth}
    \centering
        \includegraphics[height=4.5cm]{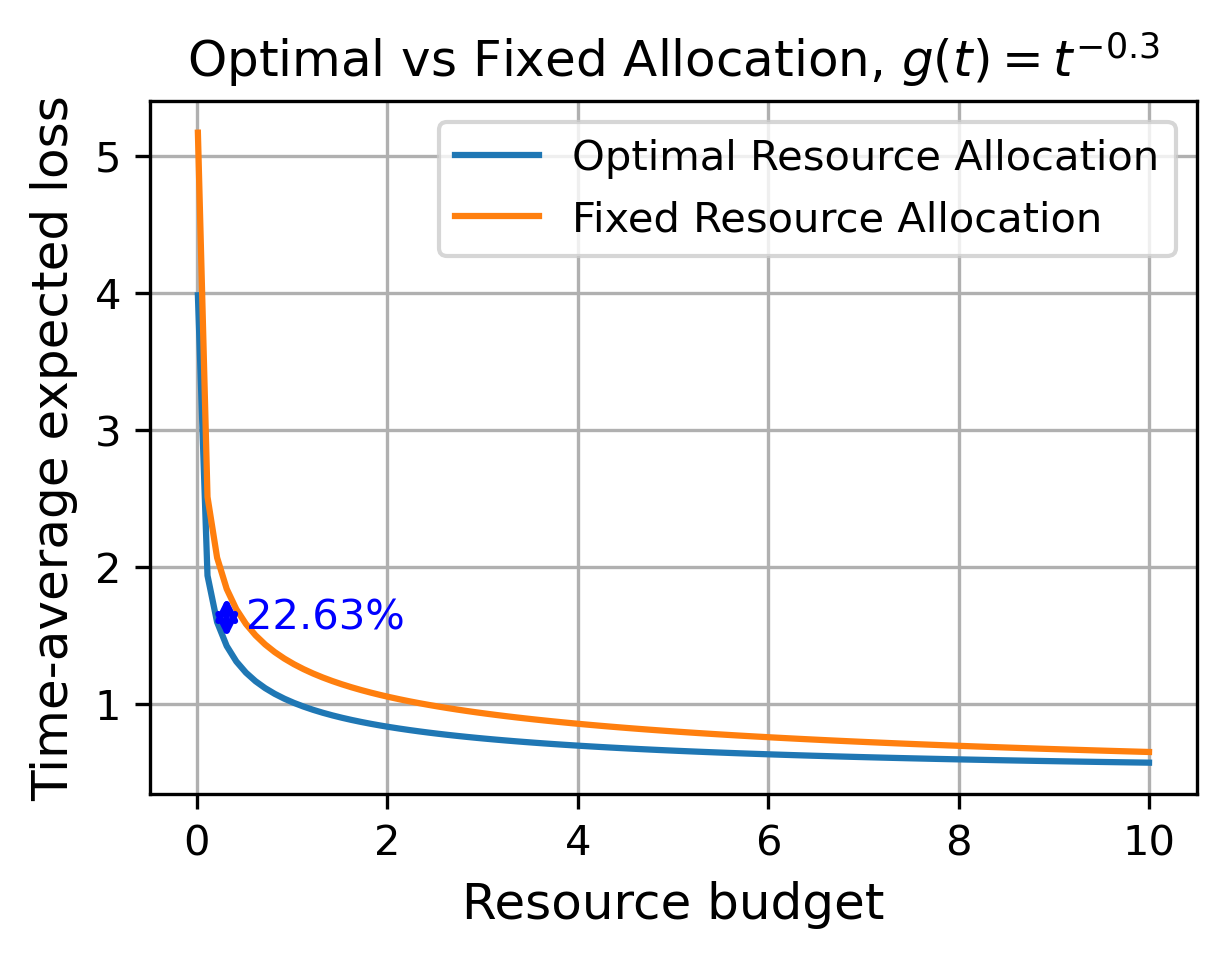}
        \caption{Comparison between constraint-binding optimal and fixed resource allocation policies. $Y\sim \text{Exp}(1)$, $\bar{g}(t)=t^{-0.3}$, $ \sigma_e=1,M=20$.}
        \label{fig:add_sim_sqrt}
    \end{subfigure}
    \hfill
    \begin{subfigure}{0.48\linewidth}
    \centering
        \includegraphics[height=4.5cm]{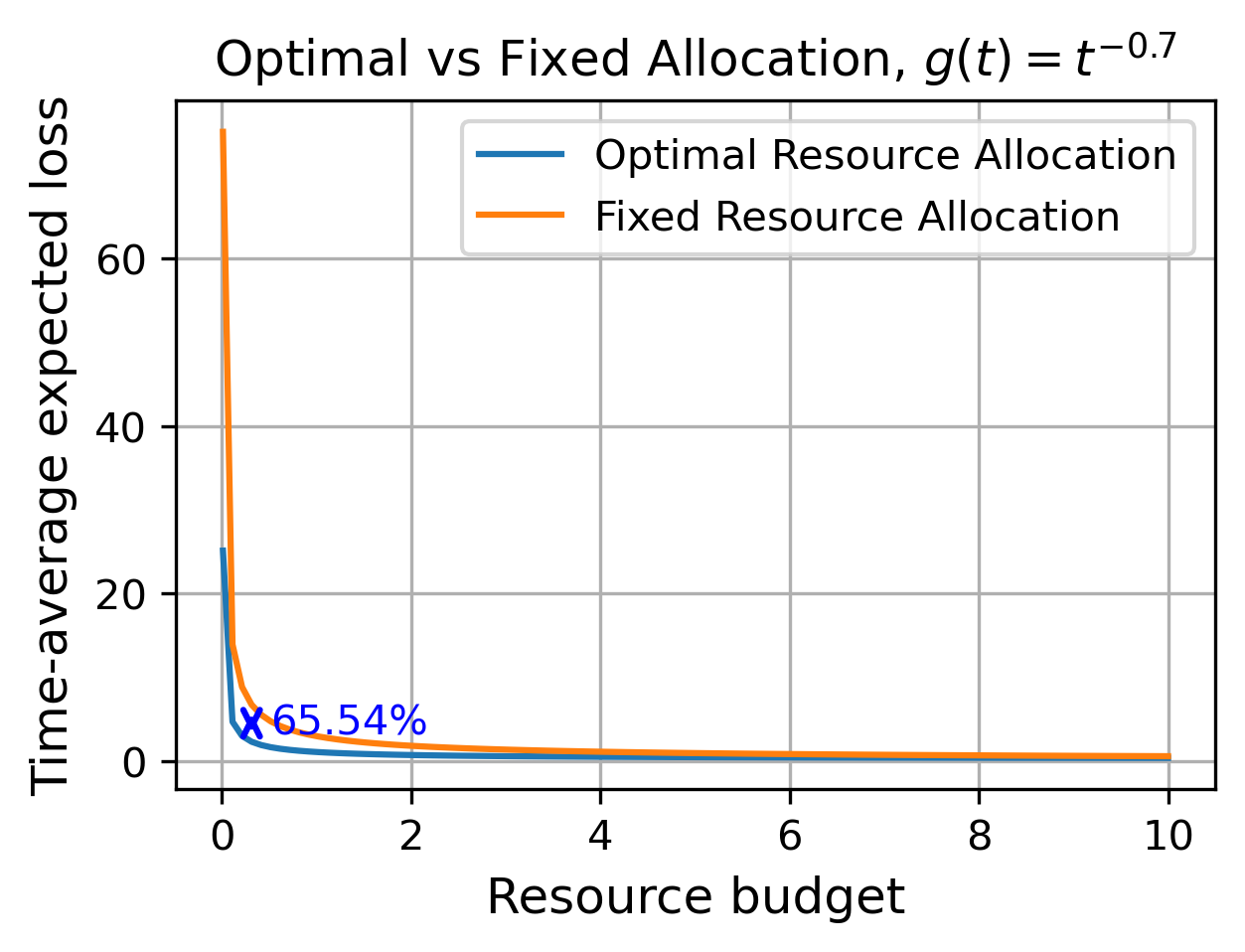}
        \caption{Comparison between constraint-binding optimal and fixed resource allocation policies. $Y\sim \text{Exp}(1)$, $\bar{g}(t)=t^{-0.7}$, $ \sigma_e=1,M=20$.}
        \label{fig:add_sim_sqrt}
    \end{subfigure}
    \caption{
    Comparison of optimal and fixed resource allocation policies under sudden concept drift for different expected loss curves.
    All simulations were executed on a MacBook Air~M3 with 16\,GB of RAM.
    }
\end{figure}

The simulation results demonstrate that the improvement over fixed resource allocation is greater when the expected loss curve decreases more rapidly in the initial phase, as front-loading policy effectively utilizes resources at the beginning of a concept.

We emphasize that the observed performance gains depend critically on the available resource budget. When the budget for model updates is very limited, improvements are naturally constrained, whereas with an abundant budget the different deployment policies converge to similar performance. The results shown in the figures correspond to intermediate budgets, where gains are most pronounced. These findings -- both qualitative and quantitative -- offer valuable guidance for selecting training and deployment strategies according to the provider’s cost sensitivity and available computational resources.

\end{document}